\documentclass[10pt,twocolumn,letterpaper]{article}

\usepackage{iccv}
\usepackage{times}
\usepackage{epsfig}
\usepackage{graphicx}
\usepackage{amsmath}
\usepackage{amssymb}
\usepackage{dsfont}
\usepackage{siunitx}
\usepackage{subfigure}
\usepackage{multirow}
\usepackage{caption}
\usepackage{booktabs}       %
\usepackage{standalone}
\usepackage{xspace}

\graphicspath{{./images_lowres/}}

\def\argmax{\mathop{\rm argmax}}

\def\1{\mathds{1}}

 \makeatletter
 \DeclareRobustCommand\onedot{\futurelet\@let@token\@onedot}
 \def\@onedot{\ifx\@let@token.\else.\null\fi\xspace}
 \def\eg{e.g\onedot} 
 \def\ie{i.e\onedot}

 \makeatother

\newcommand{\myparagraph}[1]{\paragraph{#1}}

\usepackage[breaklinks=true,bookmarks=false]{hyperref}

\iccvfinalcopy %

\def\iccvPaperID{1496} %

\ificcvfinal\fi

\begin{document}

\title{Attentive Explanations: Justifying Decisions and Pointing to the Evidence}

 \newcommand{\authSpace}{&}%
 \author{
 \begin{tabular}{ccc}
 Dong Huk Park$^{1}$ \authSpace Lisa Anne Hendricks$^{1}$ \authSpace Zeynep Akata$^{1,2}$ \\ 
  Bernt Schiele$^{2}$ \authSpace Trevor Darrell$^{1}$ \authSpace Marcus Rohrbach$^{1}$\vspace{2mm} \\ 
  \multicolumn{3}{c}{$^{1}$UC Berkeley EECS,  CA, United States}\\%  \ \ \ \ \ \ \ \ \ \ \ \  $^{3}$Facebook AI Research}\\
 \end{tabular} \\
 \begin{tabular}{ccc}
 \multicolumn{3}{c}{$^{2}$Max Planck Institute for Informatics, Saarland Informatics Campus, Saarbr\"ucken, Germany}\\
 \end{tabular} 
 }

\newcommand{\toAdd}[1]{\textcolor{red}{\textbf{Todo #1}}}
\maketitle
\begin{abstract}
Deep models are the defacto standard in visual decision problems due to their impressive performance on a wide array of visual tasks. However, they are frequently seen as opaque and are unable to explain their decisions. In contrast, humans can justify their decisions with natural language and point to the evidence in the visual world which supports their decisions. We propose a method which incorporates a novel explanation attention mechanism; our model is trained using textual rationals, and infers latent attention to visually ground explanations. We collect two novel datasets in domains where it is interesting and challenging to explain decisions. First, we extend the visual question answering task to not only provide an answer but also visual and natural language explanations for the answer. Second, we focus on explaining human activities in a contemporary activity recognition dataset. We extensively evaluate our model, both on the justification and pointing tasks, by comparing it to prior models and ablations using both automatic and human evaluations. 
\end{abstract}

\section{Introduction}
\label{sec:intro}

Humans are surprisingly good at explaining their decisions, even though their explanations do not necessarily align with their initial reasoning \cite{wick1992reconstructive}. %
Still, explaining decisions is an integral part of human communication, understanding, and learning.
Therefore, we aim to build models that explain their decisions, something which comes naturally to humans. %
Explanations can take many forms.  
For example, humans can explain their decisions with natural language, or by pointing to visual evidence. 

\begin{figure}[t]
\small 
\includegraphics[width=\linewidth]{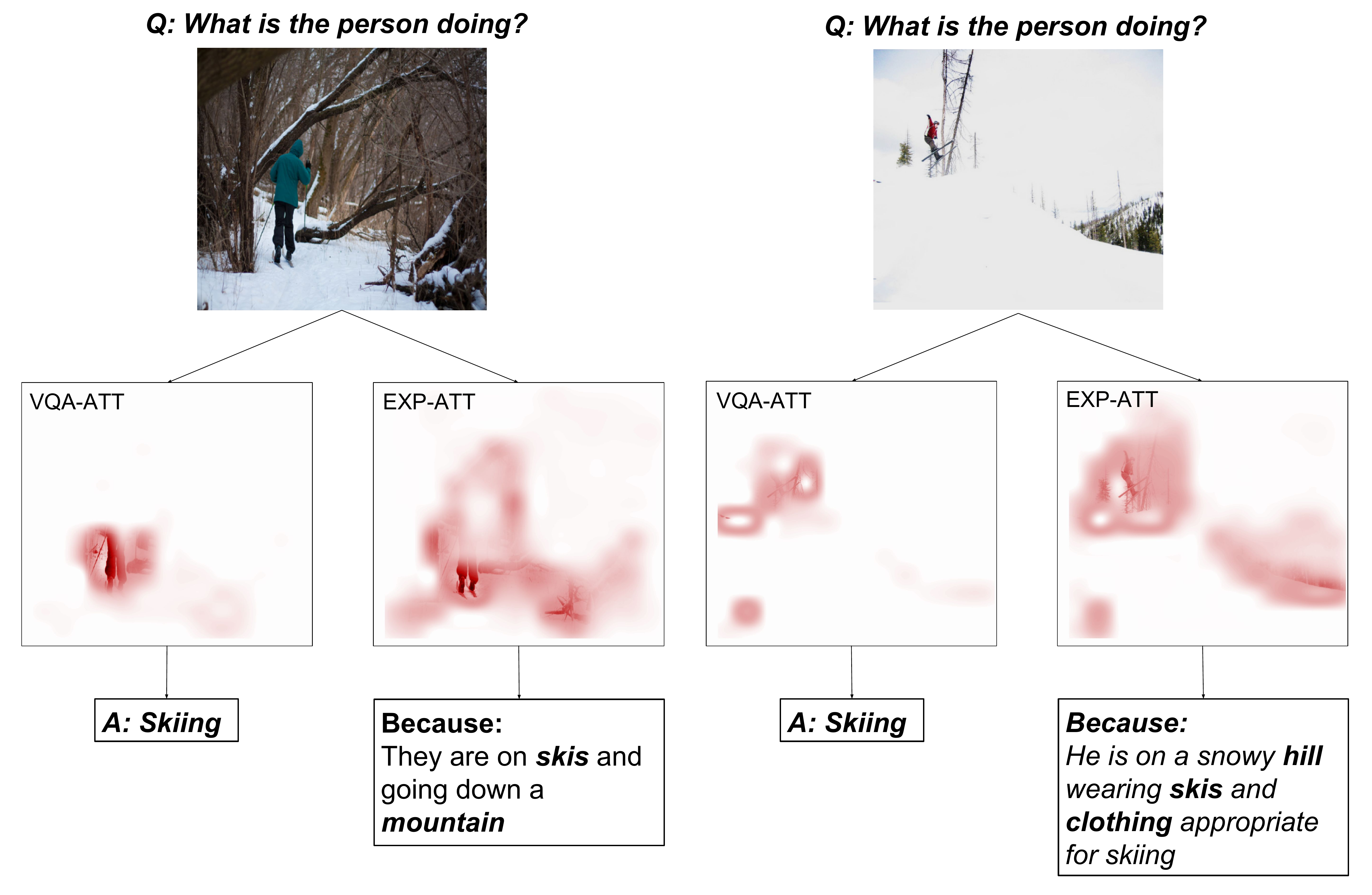}
\caption{\emph{Attentive Explanations}: For a given question and corresponding image, we predict the answer and explain it by generating a natural language justification and introspecting the model with two attention mechanisms, the first for the answer (VQA-ATT) and the second for the justification (EXP-ATT).} %
\label{fig:teaser}
\end{figure}

We show here that deep models can demonstrate similar competence, and develop a novel multi-modal model which textually justifies decisions and visually grounds evidence via two attention mechanisms. Previous methods were able to  provide a text-only explanation conditioned on an image in context of a task, or were able to visualize active intermediate units in a deep network performing a certain task, but were unable to provide explanatory text grounded in an image. In contrast, our  Pointing and Justification-based explanation (PJ-X) model is  explicitly multi-modal, not only generating textual justifications but also providing two visual attentions for decision and justification, respectively (see \autoref{fig:teaser}).

Generating convincing explanations calls for models to not only recognize objects, activities, and attributes, but to highlight which visual elements are important for a classification decision.
To produce convincing explanations, we propose a \textit{multi-modal} explanation system which provides explanations both verbally and by pointing. To illustrate, consider the two images in \autoref{fig:teaser}. In both examples, the question ``What is the person doing?'' is asked, and the model correctly answers ``Skiing.''
Though both images share common visual elements (\eg, skis and snow), the textual justifications reflect differences in the two images: while one justifies the answer ``Skiing'' by discussing skis and \emph{mountain}, the other justifies the answer with skis, \emph{hill}, and \emph{clothing}.
With respect to pointing, in both examples, the VQA-ATT attention map (left), which is generated as the model makes its decision, focuses on the skis and the legs, revealing what visual cue the model relies on when answering the question. However, the EXP-ATT map (right), which is generated when explaining the decision it has made, points to different evidence discussed by the textual justifications.  This demonstrates that the model need not attend to the same evidence when making a decision and subsequently justifying its decision.
The EXP-ATT map allows us to confirm whether the model is actually attending to the discussed items when generating the textual justification (as opposed to just memorizing justification text), and by comparing it to the VQA-ATT map, we can determine if the model attends to the same regions when making a decision as it does when explaining its decision.
Following  \cite{biran2014justification} and \cite{hendricks16eccv} we differentiate between \textit{introspective} explanations which reflect the decision process of a network (e.g., ``The model decided this person is skiing because it focused on this region when making its decision'') and \textit{justification} explanations which discuss evidence that supports a decision (\eg, ``This person is skiing because he is on a snowy hill wearing skis'') without necessarily reflecting a neural network decision process but reflecting explanations given by humans.
Introspective models can lead to better understanding of network decision processes, but justification systems can potentially be clearer to end-users who are not familiar with deep models.  
The PJ-X model encompasses both philosophies.
Whereas text generated by the PJ-X model may not directly reflect the model's decision process, it can provide straightforward explanations which are easy to understand by end-users.
By including attention activations used during the decision and justification processes, PJ-X is also introspective.

Introspective explanation models illuminate the underlying mechanism of a model's decision.
Thus, to develop introspective explanation models, a researcher only needs access to data and the model itself.
In contrast, justification explanation systems aim to discuss evidence which supports a decision in a human understandable format.
Thus, we believe it is important to have access to ground truth human justifications for evaluation of justification systems.
There is a dearth of datasets which include examples of how humans justify specific decisions.
We propose and collect complementary explanation datasets for two challenging vision problems: activity recognition and visual question answering (VQA).
We collect both training and evaluation data for textual justifications as well as evaluation data for the pointing task.
In sum, we present a model which goes beyond current visual explanation systems by producing multi-modal, grounded explanations. We incorporate a novel explanatory attention step in our method, which allows it to visually ground explanation text. In order to generate satisfactory explanations, we collect two new datasets which include human explanations for both activity recognition and visual question answering.  
Our proposed Pointing and Justification Explanation (PJ-X) model outperforms strong baselines. We additionally show that our VQA part of the model improves slightly over MCB \cite{fukui16emnlp}, the VQA 2016 challenge winner, and is more efficient to train and test.

\section{Related Work}
\label{sec:related}

\myparagraph{Explanations.}  Early textual explanation models span a variety of applications (e.g., medical \cite{shortliffe1975model} and feedback for teaching programs \cite{lane2005explainable, van2004explainable, core2006building}) and are generally template based.  More recently, \cite{hendricks16eccv} developed a deep network to generate natural language justifications of a fine-grained object classifier. 
However, unlike our model, it does not provide multi-modal explanations and the model is trained on descriptions rather than reference explanations.

A variety of work has proposed methods to visually explain decisions.  Some methods find discriminative visual patches  \cite{doersch2012makes,berg2013you} whereas others aim to understand intermediate features which are important for end decisions~\cite{zeiler2014visualizing,escorcia2015relationship,zhou2014object} \eg what does a certain neuron represent.
PJ-X points to visual evidence via an attention mechanism which is an intuitive way to convey knowledge about what is important to the network without requiring domain knowledge.
In contrast to previous work, PJ-X generates \emph{multi-modal explanations} in the form of explanatory sentences and attention maps pointing to the visual evidence.
As discussed in Section 1 explanation systems can either be \textit{introspective} systems or \textit{justification} systems.
In this paradigm, models like  \cite{hendricks16eccv} which highlight discriminative image attributes without access to a specific model are considered justification explanations, whereas models like~\cite{zeiler2014visualizing} which aim to illuminate the inner workings of deep networks are considered introspective explanations.
We argue that both are useful; justifications can provide helpful information for humans in an easily digestible format, whereas introspective explanations can provide insight into a model's decision process, though it may be harder for a human unfamiliar with deep learning to understand.
Our model strives to satisfy both definitions; providing textual explanations fits the definition of justification explanations whereas visualizing where the system attends provides introspective explanations.

\begin{figure*}[t]
  \includegraphics[width=\linewidth]{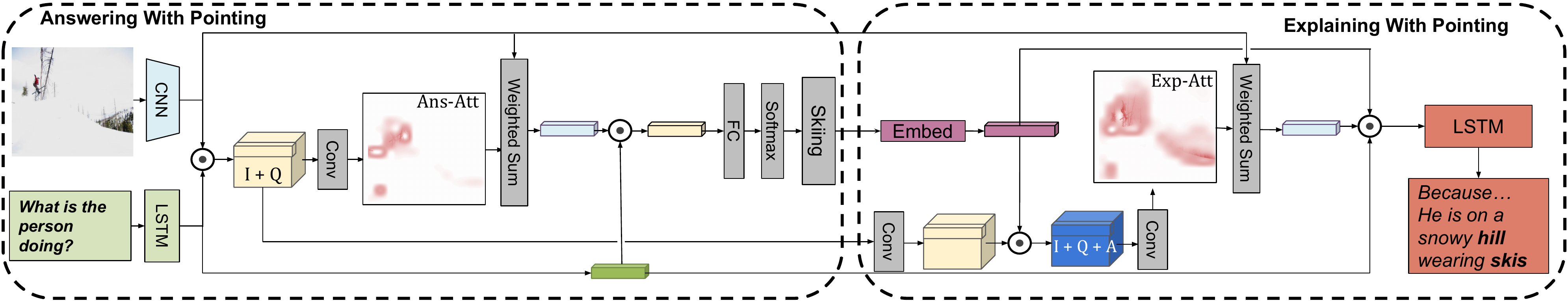}
\caption{Our Pointing and Justification (PJ-X)
architecture generates a multi-modal explanation which includes a textual justification (``He is on a snowy hill wearing skis'') and points to the visual evidence.  Our model consists of two ``pointing'' mechanisms: answering with pointing (left) and explaining with pointing (right).}
\label{fig:attentive_explanation}
\end{figure*}

\myparagraph{Visual Question Answering and Attention.}
Initial approaches to VQA used full-frame representations \cite{malinowski15iccv}, but most recent approaches use some form of spatial attention~\cite{yang2015stacked,xu2015ask,zhu16cvpr,chen2015abc,xiong16dynamic,shih2015look,fukui16emnlp,DBLP:journals/corr/KimOKHZ16}. We base our method on \cite{fukui16emnlp}, \ie the winner of VQA 2016 challenge, and predict a latent weighting (attention) of spatially localized image features based on the question, however we use an element-wise product as opposed to compact bilinear pooling.
Concurrent work \cite{DBLP:journals/corr/KimOKHZ16} has explored the element-wise product for VQA just as we do in our method, however~\cite{DBLP:journals/corr/KimOKHZ16} improves performance by applying hyperbolic tangent (TanH) after the multi-modal pooling whereas we improve by applying signed square-root and L2 normalization. 

\myparagraph{Activity Recognition.}
Recent work on activity recognition in still images relies on a variety of cues, such as pose and global context \cite{pishchulin14gcpr,mallya16eccv}.
However, although cues like pose may influence model performance, activity recognition models are not capable of indicating which factors influence a decision process.
In contrast, explanations aim to reveal which parts of an image are important for classification.

\section{Pointing and Justification Model (PJ-X)}
\label{sec:method}

The goal of our work is to justify why a decision was made with natural language, and point to the evidence for both the decision and the textual justification provided by the model.
We deliberately design our Pointing and Justification Model (PJ-X) to allow training these two tasks as well as the decision process jointly.
Specifically we want to rely on natural language justifications and the classification labels as the only supervision. %
We design the model to learn how to point in a latent way. For the pointing we rely on an attention mechanism \cite{bahdanau2014neural} which allows the model to focus on a spatial subset of the visual representation. As the model ignores all spatial visual features it does not (or insignificantly) attend to, this pointing also allows us to introspect the model. Our model uses two different attentions, one makes predictions and another generates explanations.

We first predict the answer given an image and question. Then given the answer, question, and image, we generate the textual justification. In both cases we include a latent attention mechanism which allows to introspect where the question or the answer points to.
An overview of our double attention model is presented in \autoref{fig:attentive_explanation}.

\myparagraph{Learning to answer.} In visual question answering the goal is to predict an answer given a question and an image. For activity recognition we do not have an explicit question. %
 Thus, we ignore the question which is equivalent to setting the question representation to $f^Q(Q)=1$, a vector of ones. %

To be able to introspect the answering process we want the model to select the area of the image which gives the evidence for the answer. This can be achieved using an attention model. While we rely on the overall architecture from the state-of-the-art MCB attention model~\cite{fukui16emnlp}, we remove the core contribution of \cite{fukui16emnlp}, the MCB unit to pool multi-modal features. Instead we propose to use the simpler element-wise multiplication $\odot$ for pooling after a fully-connected layer for embedding the visual feature which learns an alignment between between the visual and textual representation. We found that this leads to similar performance, but much faster training. Comparison on the VQA dataset~\cite{antol2015vqa} between our model and the state-of-the-art model can be found in \autoref{sec:vqa_model}.

In detail, we extract  spatial image features $f^{I}(I,n,m)$ from the last convolutional layer of ResNet-152 followed by $1\times1$ convolutions ($\bar{f}^{I}$) giving a $2048\times N\times M$ spatial image feature. We encode the question $Q$ with a 2-layer $LSTM$, which we refer to as $f^Q(Q)$. We combine this and the spatial image feature using element-wise multiplication followed by signed square-root, L2 normalization, and Dropout, and two more layers of $1\times1$ convolutions with ReLU in between, which operate on the spatial feature map location $n\in N$ and $m\in N$:
\begin{align}
\label{eq:pointA}
\bar{f}^{IQ}(I,n,m,Q) = & (W_1 f^{I}(I,n,m) + b_1) \odot f^Q(Q)\\
f^{IQ}(I,Q) =& L2(\textit{signed\_sqrt}(\bar{f}^{IQ}(I,Q)))\\
\bar{\alpha}_{n,m}^{pointA} = &f^{pointA}(I,n,m,Q)\\ 
= & W_{3}\rho(W_{2} f^{IQ}(I,Q) + b_{2}) + b_{3}
\end{align}
with ReLU $\rho(x) = max(x,0)$.
This process gives us a $N\times M$ attention map $\bar{\alpha}_{n,m}$. We apply softmax to produce a normalized soft attention map, which thus points at the evidence of the answer ($pointA$):
\begin{equation}
\label{eq:attention}
\alpha_{n,m}^{pointA} = \frac{\exp(\bar{\alpha}_{n,m}^{pointA})}{\sum_{i=1}^N\sum_{j=1}^M{\exp(\bar{\alpha}_{i,j}^{pointA})}}
\end{equation}

The attention map is then used to take the weighted sum over the image features and this representation is once again combined with the LSTM feature to predict the answer $\hat{y}$ as a classification problem over all answers $Y$. 
\begin{align}
\bar{f}^{y}(I,Q) = & (\sum_{x=1}^N\sum_{y=1}^M{\alpha_{n,m}^{pointA}f^{I}(I,n,m) }) \odot f^Q(Q) \\
{f}^{y}(I,Q) = & W_4 \bar{f}^{y}(I,Q) + b_4\\
p(y|I,Q) = & Softmax({f}^{y}(I,Q))\\
\label{eq:vqa_prediction}
\hat{y}=&\argmax_{y\in Y}p(y|I,Q)
\end{align}

\paragraph{Learning to justify.}
We argue that to generate  a textual justification for VQA, we should condition it on the question, the answer, and the image. For instance, to be able to explain ``Because they are Vancouver police'' in~\autoref{fig:VQA_examples}, the model needs to see the question, \ie ``Can these people arrest someone?'', the answer, \ie ``Yes'' and the image, \ie the ``Vancouver police'' banner on the motorcycles. 

We model this by first using a second attention mechanism and then using the localized feature as input to an LSTM which generates the explanations. In this way we hope to uncover which parts of the image contain the evidence for the justification.

More specifically, the answer predictions are embedded in a $d$-dimensional space followed by $\tanh$ non-linearity and a fully connected layer:
\begin{align}
f^{yEmbed}(\hat{y}) = & W_6(tanh(W_5 \hat{y} + b_5))+b_6
\end{align}

To allow the model to learn how to attend to relevant spatial location based on the answer, image, and question, we combine this answer feature with Question-Image embedding $\bar{f}^{IQ}(I,Q)$. After applying $1\times1$ convolutions, element-wise multiplication followed by signed square-root, L2 normalization, and Dropout, the resulting multimodal feature is flattened to a $14\times14$ attention map similarly as the previous attention step: %
\begin{align}
\bar{f}^{IQA}(I,n,m,Q,\hat{y}) = & (W_7 \bar{f}^{IQ}(I,Q,n,m) + b_7) \\
&\odot f^{yEmbed}(\hat{y}))\\
f^{IQA}(I,Q,\hat{y}) =& L2(\textit{signed\_sqrt}(\bar{f}^{IQA}(I,Q,\hat{y})))\\
\bar{\alpha}_{n,m}^{pointX} = &f^{pointX}(I,n,m,Q,\hat{y})\\ 
= & W_{9}\rho(W_{8} f^{IQA}(I,Q,\hat{y}) + b_{8}) + b_{9}
\end{align}
This process gives us a $N\times M$ attention map $\bar{\alpha}_{n,m}$. We apply softmax to produce a normalized soft attention map, which aims to point at the evidence of the generated explanation ($pointX$):
\begin{equation}
\label{eq:attention}
\alpha_{n,m}^{pointX} = \frac{\exp(\bar{\alpha}_{n,m}^{pointX})}{\sum_{i=1}^N\sum_{j=1}^M{\exp(\bar{\alpha}_{n,m}^{pointX})}}
\end{equation}

Using this second attention map, we compute the attended visual representation, and merge it with the LSTM feature that encodes the question and the embedding feature that encodes the answer:
\begin{align}
\
f^{X}(I,Q,\hat{y}) = & (W_{10} \sum_{x=1}^N\sum_{y=1}^M{\alpha_{n,m}^{pointX}f^{I}(I,n,m) } + b_{10})\\
& \odot (W_{11} f^Q(Q) + b_{11}) \\
& \odot f^{yEmbed}(\hat{y})
\end{align}

This combined feature is then fed into an LSTM decoder to generate explanations that are conditioned on image, question, and answer.

It predicts one word $w_{t}$ at each time step $t$  conditioned on the previous word and the hidden state of the LSTM:
\begin{align}
h_{t} = f^{LSTM}(f^{X}(I,Q,\hat{y}),w_{t-1},h_{t-1})\\
w_{t} = f^{pred}( h_{t}) = Softmax(W_{pred} h_{t} + b_{pred}  )
\end{align}

{
\renewcommand{\arraystretch}{1.2}

\begin{table*}[t]
\small
\center
  \begin{tabular}{@{}l@{}r@{\ \ }r@{\ \ \ }r@{\ }l@{}r@{\ }l@{ \ }r@{\ }r@{\ }} 
    \textbf{Dataset} & \#imgs & \#classes & Desc.&(\#w) & Expl.&(\#w) & \#att & maps\\     \hline
    CUB~\cite{CaltechUCSDBirdsDataset,RALS16} & $11$k & $200$ & $58$k&($17$) & 0 & & & 0 \\
\hline
MSCOCO~\cite{lin14eccv},VQA \cite{antol2015vqa} & $123$k & $\geq 3000$ & $616$k&($10.6$) & 0 & & & 0  \\
VQA-X (ours) & $20$k & $3000$ & 0 && $30$k&(8.1) & & $1500$ \\%$102$k&(10.6)
\hline
    MHP~\cite{APGS14,pishchulin14gcpr,RAMTSL16} & $25$k & $410$ & $75$k&($15$) & 0 & & & 0 \\
ACT-X (ours) & $18$k & 397& 0 &&$54$k&($13$) & & $1500$ \\  
\bottomrule
  \end{tabular}
\caption{Statistics of datasets. Desc.=Descriptions, Expl.=Explanations, \#w=average number of words, \#att maps=number of attention map annotations.}
\label{tab:datasets}
\end{table*}
}

\section{Visual Explanation Datasets}
\label{sec:datasets}

We propose two explanation datasets: Visual Question Answering Explanation (VQA-X) and MPI Human Pose Activity Explanation (ACT-X). A summary of dataset statistics is presented in~\autoref{tab:datasets}.

\begin{figure}[t]
\center
  \includegraphics[width=\linewidth]{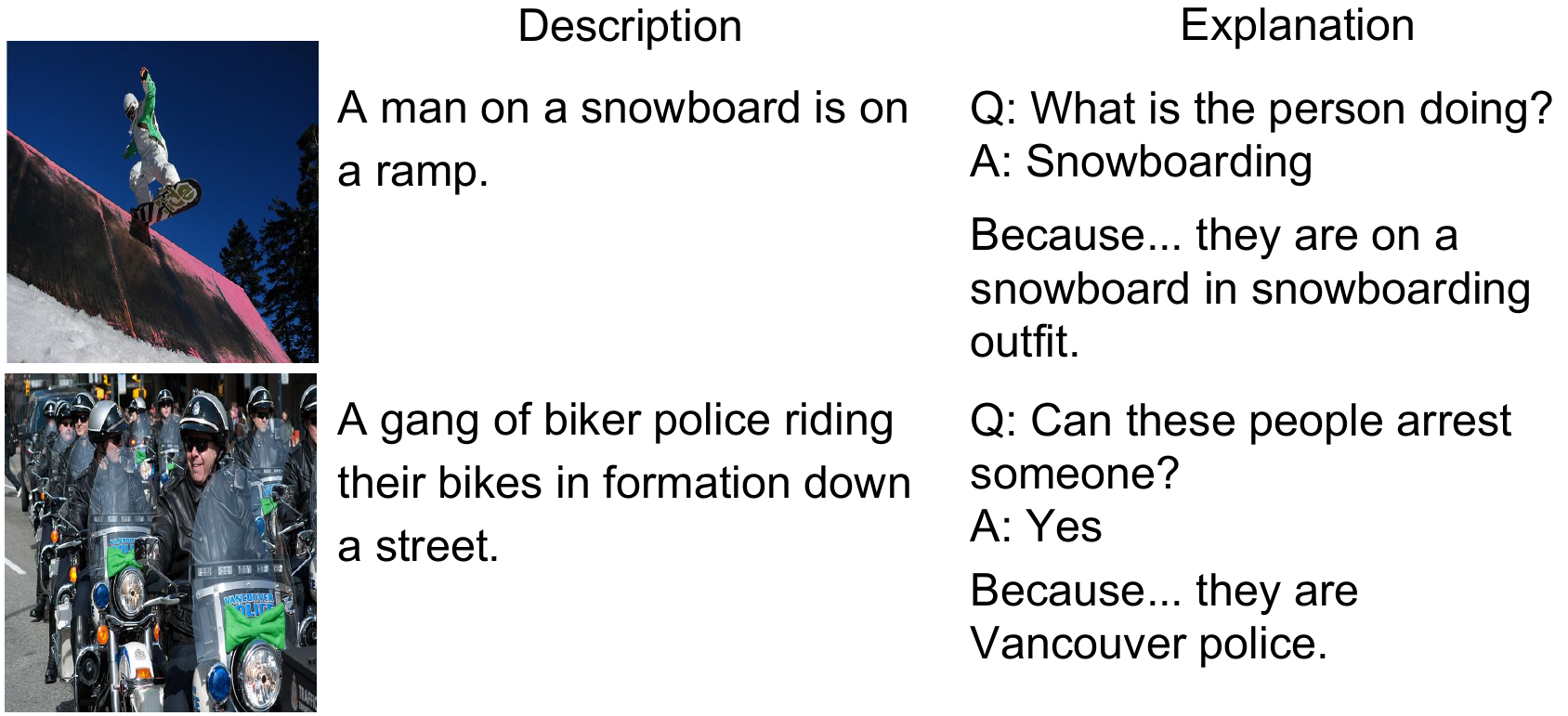}
\caption{In comparison to the descriptions, our explanations focus on the visual evidence that pertains to the question and answer instead of generally describing objects in the scene.}
\label{fig:VQA_examples}
\end{figure}

\myparagraph{VQA Explanation Dataset (VQA-X).} The Visual Question Answering (VQA) dataset~\cite{antol2015vqa} contains open-ended questions about images which require understanding vision, natural language, and commonsense knowledge to answer. The dataset consists of approximately $200$K MSCOCO images~\cite{lin2014microsoft}, with $3$ questions per image and $10$ answers per question. %
We select $18,357$ images with $20$K question/answer (QA) pairs from the VQA training set and $2$K QA pairs (991 images) from the VQA validation set, which are later divided into $1$K QA pairs each for validation and testing. 
The QA pairs were selected based on a few simple heuristics that would remove pairs that require trivial explanations, such as Q: ``What is the color of the banana?'' etc. We collected $1$ explanation per data point for the training set and $5$ explanations per data point for the validation and test sets. The annotators were asked to provide a proper sentence or clause that would come after the proposition ``because'' as explanations to the provided image, question, and answer triplet. Examples for both descriptions, \ie from MSCOCO dataset, and our explanations are presented in \autoref{fig:VQA_examples}. 

\begin{figure}[t]
\center
  \includegraphics[width=\linewidth]{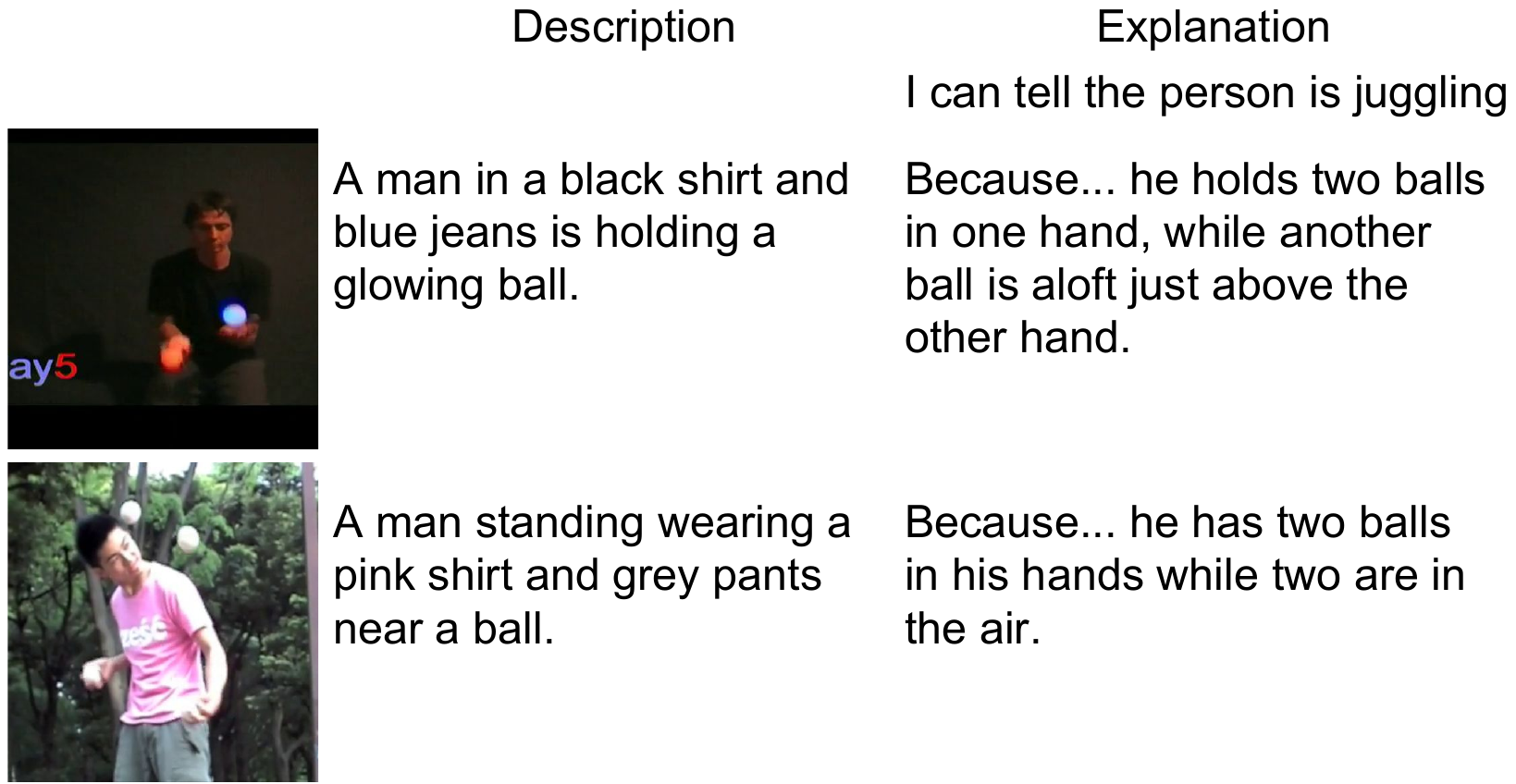}
\caption{Our ACT-X dataset contains images from MHP~\cite{APGS14} dataset and our activity explanations. For MHP, \cite{RAMTSL16} collected one-sentence descriptions. Our explanations are task specific whereas descriptions are more generic.}
\label{fig:MHP_examples}
\end{figure}

\myparagraph{Action Explanation Dataset (ACT-X).} The MPI Human Pose (MHP) dataset~\cite{APGS14} contains $25$K images extracted from videos downloaded from Youtube. From the MHP dataset, we selected all images that pertain to $397$ activities, resulting in $18,030$ images total (3 splits with training set having 12,607 images, the validation set with 1,802 images, and finally the test set with 3,621 images).
For each image we collected $3$ explanations. During data annotation, we asked the annotators to complete the sentence ``I can tell the person is doing (action) because..''  where the action is the ground truth activity label. We also asked them to use at least 10 words and avoid mentioning the activity class in the sentence. %
MHP dataset also comes with $3$ sentence descriptions provided by~\cite{RAMTSL16}. Some examples of descriptions and explanations can be seen in \autoref{fig:MHP_examples}. 

\begin{figure*}
\centering
\subfigure[VQA-X]{\label{fig:vqa_seg_gt}\includegraphics[width=80mm, height=60mm]{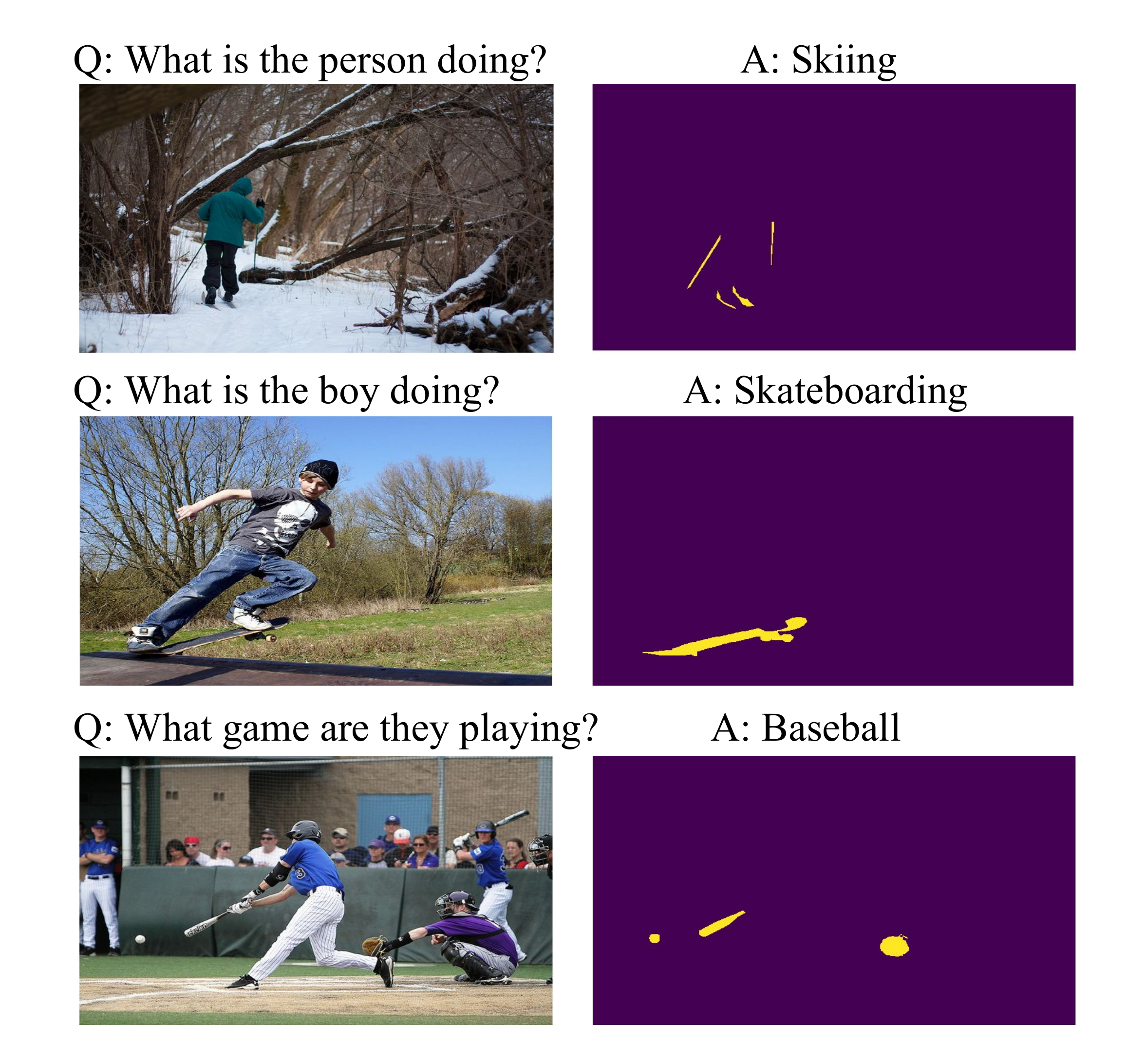}}
\subfigure[ACT-X]{\label{fig:act_seg_gt}\includegraphics[width=80mm, height=60mm]{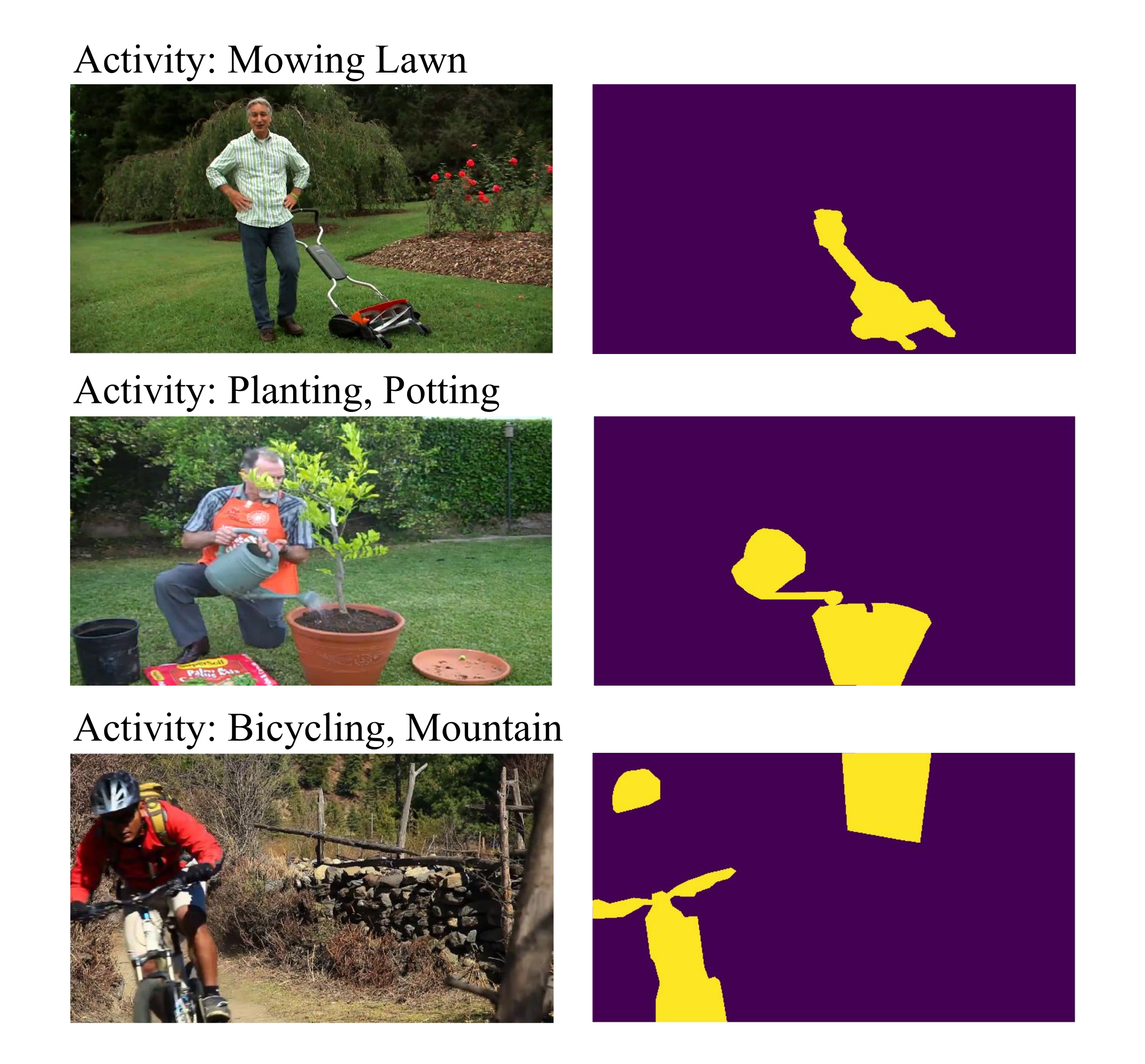}}
\caption{Human-Annotated Attention Maps. Figure on the left: Example annotations collected on VQA-X dataset for the pointing task. Figure on the right: Example annotations collected on ACT-X dataset for the pointing task. In both cases, the visual evidence that justifies the answer is segmented in yellow.}
\label{fig:GroundTruthMap}
\end{figure*}

\myparagraph{Ground truth for pointing.} In addition to textual justification, we collect attention maps from humans for both VQA-X and ACT-X datasets in order to evaluate if the attention of our model corresponds to where humans think the evidence for the answer is. Human-annotated attention maps are collected via
Amazon Mechanical Turk where we use the segmentation UI interface from the OpenSurfaces Project \cite{bell13opensurfaces}. Annotators are provided with an image and an answer (question and answer pair for VQA-X, class label for ACT-X). They are asked to segment objects and/or regions that most prominently justify the answer. For each dataset we randomly sample 500 images from the test split, and for each image we collect 3 attention maps. The collected annotations are used for computing the Earth Mover's Distance to evaluate attention maps of our model against several baselines. Some examples can be seen in \autoref{fig:GroundTruthMap}.

\section{Experiments}
\label{sec:exp}
In this section, after detailing the experimental setup, we present our model for visual question answering, our results for textual justification and visual pointing tasks. Finally, we provide and analyze qualitative results for both tasks.

\subsection{Experimental Setup}
\label{sec:experimentalsetup}
Here, we detail our experimental setup in terms of model training, hyperparameter setting and evaluation metrics.

\myparagraph{Model training and hyperparameters.}
For VQA, our model is pre-trained on the VQA training set \cite{antol2015vqa} to achieve state-of-the-art performance on predicting answers, but we either freeze or finetune the weights of the prediction model when training on explanations as the VQA-X dataset is significantly smaller than the original VQA training set. We refer the finetuned model as `Findtuned' throughout the paper and all other VQA models have their weights fixed. For activity recognition prediction and explanation components of the pipeline is trained jointly. 
The spatial feature size of our model is $N=M=14$. 
For VQA, we classify with the $3000$ most frequently occurring answers on the training set (\ie $|Y|=3000$) whereas for activity recognition, $|Y|=397$. We set the answer embedding size as $d=300$ for both tasks.
We train all our models on the training set, set hyperparameters on the validation set, and report results on the test set, the  splits are detailed in \autoref{sec:datasets}. 

\myparagraph{Evaluation metrics.} We evaluate our textual results w.r.t BLEU-4~\cite{papineni2002bleu}, METEOR~\cite{banerjee2005meteor}, ROUGE~\cite{L04}, CIDEr~\cite{vedantam2015cider} and SPICE~\cite{spice2016} metrics, based on the degree of similarity between generated and ground truth sentences.
We also include human evaluation as automatic metrics do not always reflect with human preference. We randomly choose 250 images each from the test sets of the VQA-X and ACT-X datasets and then ask 3 humans for each image to judge whether a generated explanation is better than, worse than, or equivalent to a ground truth explanation (we note that human judges do not know what explanation is ground truth and the order is randomized).
We report the percentage of generated explanations which are equivalent to or better than ground truth human explanations, when at least 2 out of 3 human judges agree.

For visual pointing task, we use Earth Mover's Distance (EMD)~\cite{RubnerTG98iccv}, which measures the distance between two probability distributions over a region, and rank correlation, which was used in~\cite{DBLP:journals/corr/DasAZPB16}, as our evaluation metrics. EMD reflects the minimum amount of work that must be performed to transform one distribution into the other by moving ``distribution mass''. EMD captures the notion of distance between two sets or distributions instead of two single points. We use the code from \cite{pele2009} to compute EMD. For computing rank correlation, we follow~\cite{DBLP:journals/corr/DasAZPB16} where we scale our attention maps and the human attention maps from the VQA-HAT dataset to $14 \times 14$, rank the pixel values, and then compute correlation between these two ranked lists.

\subsection{Visual Question Answering Model}
\label{sec:vqa_model}
The VQA model that we use throughout the experiments is based on the state-of-the-art MCB model~\cite{fukui16emnlp}, but trains and evaluates faster (reduction of $\sim 30\%$). The main difference between the two models is how they combine two different representations and create multimodal features. We evaluate our VQA model using the same accuracy measure as in the VQA challenge.

Instead of doing Compact Bilinear Pooling \cite{gao16cvpr} between the two representations, our model simply embeds the encoded image feature using $1\times1$ convolutions and applies element-wise multiplication between the image embedding and the LSTM feature. While the MCB model aims to create a rich multimodal feature by approximating the outer product of two representations, our model tries to learn the proper alignment between features so that when merged with element-wise multiplication, it creates a feature that is as powerful as the MCB feature. Similar to \cite{fukui16emnlp}, the merged representation is normalized by applying signed square root and L2-normalization. As shown in \autoref{tab:VQA}, our VQA model leads to a moderate 0.5\% improvement on the training set and 0.6\% on train-val set, though $\sim 30\%$ faster.

\begin{table}[tb]
\centering
\small
\begin{tabular}{l|rr}
\toprule
 \bf  &  \multicolumn{2}{c}{Training data} \\ 
 \bf Method &  Train & Train+Val \\ \cmidrule(lr){1-1}\cmidrule(lr){2-3}
MCB \cite{fukui16emnlp} & 62.5 & 64.2 \\
 Our VQA model & \bf{63.0} & \bf{64.8}\\
 \bottomrule
\end{tabular}
\caption{OpenEnded results on VQA dataset \cite{antol2015vqa}, test-dev. The columns indicate the accuracy of the model after being trained on training set and train+val set, respectively. Our model achieves slightly  higher accuracy than the previous VQA challenge winner MCB \cite{fukui16emnlp} while being faster at train and test time.}
\label{tab:VQA}
\vspace{-4mm}
\end{table}

{
\setlength{\tabcolsep}{3pt}
\renewcommand{\arraystretch}{1.2}
\begin{table*}[tb]
\begin{center}
\small
\begin{tabular}{l l l l|rrrrr r|rrrrr r}
\toprule
&Train-&Att.&Answer& \multicolumn{6}{c|}{VQA-X} & \multicolumn{6}{c}{ACT-X} \\
&ing &  for & Condi-& \multicolumn{5}{c}{Automatic evaluation} & \multicolumn{1}{c|}{Human} &\multicolumn{5}{c}{Automatic evaluation} & \multicolumn{1}{c}{Human}\\
Approach &  Data & Expl. &  tioning & \multicolumn{1}{c}{B} & \multicolumn{1}{c}{M} & \multicolumn{1}{c}{R} & \multicolumn{1}{c}{C} &\multicolumn{1}{c}{S} &\multicolumn{1}{c|}{eval} & \multicolumn{1}{c}{B} & \multicolumn{1}{c}{M} & \multicolumn{1}{c}{R} & \multicolumn{1}{c}{C} &\multicolumn{1}{c}{S} &\multicolumn{1}{c}{eval}  \\
\midrule
\cite{hendricks16eccv} & Desc. & No & Yes &\multicolumn{1}{c}{--}&\multicolumn{1}{c}{--}&\multicolumn{1}{c}{--}&\multicolumn{1}{c}{--}&\multicolumn{1}{c}{--}&\multicolumn{1}{c}{--}&12.9 & 15.9  & 39.0 & 12.4 & 12.0 &7.6  \\ %
Ours on Descriptions & Desc. & Yes & Yes & 8.1 & 14.3  & 28.3  & 34.3 & 11.2 &24.0 & 6.9  & 12.9 & 28.3 & 20.3 & 7.3&18.0  \\ %
\midrule
Captioning Model & Expl. & Yes & No &17.1 &16.0 &40.4 &43.6 &7.3 &19.2 &20.7 &18.8 &44.3 &40.7 &11.3&20.4 \\ %
Ours w/o Exp-Attention & Expl. & No & Yes &25.1 &20.5 &48.7 &74.2 &11.6 &34.4& 16.9 &17.0 &42.0&33.3 &10.6 &17.6  \\ %
Ours & Expl. & Yes & Yes &25.3 &20.9  &49.8 &72.1 &12.1 & 33.6 & 24.5 &21.5 &46.9 &58.7 &16.0 & 26.4  \\ %
Ours (Finetuned) & Expl. & Yes & Yes &27.1 &20.9  &49.9 &77.2 &11.8 & -- & -- & -- & -- & -- & -- & --  \\
\bottomrule
\end{tabular}
\vspace{-2mm}
\caption{Evaluation of Textual Justifications. Evaluated automatic metrics: BLEU-4 (B), METEOR (M), ROUGE (R), CIDEr (C) and SPICE (S). Reference sentence for human and automatic evaluation is always an explanation. All in \%.  Our proposed model compares favorably to baselines.}
\vspace{-4mm}
\label{tbl:generation_automatic}
\end{center}
\end{table*}
}

\begin{table}[tb]
\small
\centering
\begin{tabular}{r|r|r|r}
\toprule
VQA-X & ACT-X & COCO Desc. & MHP Desc. \\
\midrule
$29.70$\% & $0.66$\% & $7.00$\% & $0.11$\% \\
\bottomrule
\end{tabular}

\caption{Percentage of explanations generated by the PJ-X model on the validation set which are exact copies from the training set. We evaluate on the explanation datasets (VQA-X, ACT-X) and description datasets (COCO Desc., MHP Desc.). } 
\label{tbl:sentence_overlap}
\end{table}

\subsection{Textual Justification}
\label{sec:res:textual}
We ablate our model and compare with related approaches on our VQA-X and ACT-X datasets based on automatic and human evaluation for the generated explanations.

\myparagraph{Details on compared models.} We re-implemented the state-of-the-art captioning model~\cite{donahue16pami} with an integrated attention mechanism which we refer to as ``Captioning Model''. This model only uses images and does not use class labels (i.e. the answer in VQA-X and the activity label in ACT-X) when generating textual justifications. 
We also compare with~\cite{hendricks16eccv} using publicly available code. 
For fair comparison, we use ResNet features when training \cite{hendricks16eccv} extracted from the entire image.
Generated sentences are conditioned on both the image and class predictions. \cite{hendricks16eccv} uses discriminative loss, which enforces the generated sentence to contain class-specific information, to back-propagate policy gradients when training the language generator and thus involves training a separate sentence classifier to generate rewards. Our model does not use discriminative loss/policy gradients and does not require defining a reward.
Note that \cite{hendricks16eccv} is trained with descriptions.
''Ours on Descriptions'' is another ablation in which we train the PJ-X model on descriptions instead of explanations. 
''Ours w/o Exp-Attention'' is similar to~\cite{hendricks16eccv} in the sense that there is no attention mechanism for generating explanations, however, it does not use the discriminative loss and is trained on explanations instead of descriptions. 

\myparagraph{Comparing with state-of-the-art.} Our PJ-X model performs well when compared to the state-of-the-art on both automatic evaluation metrics and human evaluations (Table \ref{tbl:generation_automatic}). ``Ours'' model significantly improves ``Ours with description'' model by a large margin on both datasets which is expected as descriptions are not collected for the task of generating explanations, it demonstrates the necessity for explanation datasets to build explanation models.
Additionally, our model outperforms \cite{hendricks16eccv} which learns to generate explanations given only description training data.  This further confirms that our new datasets with ground truth explanations are important for textual justification generation. ``Ours on Descriptions'' performs worse on certain metrics compared to~\cite{hendricks16eccv} which may be attributed to additional training signals generated from discriminative loss and policy gradients, but further investigation is due for future work.   

\myparagraph{Ablating our PJ-X model.} Comparing ``Ours'' to ``Captioning Model'' shows that conditioning explanations on a model decision is important.
Though conditioning on the answer seems to be rather helpful for ACT-X (human eval increases from 20.4 to 26.4), it seems to be essential for VQA-X (human eval increases from 19.2 to 33.6).
This is sensible because a single image in the VQA dataset can correspond to many different question and answer pairs.
Thus it is important for our model to have access to questions and answers to accurately generate the explanation.
Finally, including attention allows us to build a multi-modal explanation model.
On the ACT-X dataset, it is clear that including attention (compare ``Ours w/o Exp-Attention'' to ``Ours'') greatly improves textual justifications.
On the VQA-X dataset, ``Ours w/o Attention'' and ``Ours'' are comparable.  
Though attention does not improve scores for the textual justification task on the VQA-X dataset, it does provide us with a multi-modal explanation that provides us with added insight about a model's decision.

\myparagraph{Robustness against statistical priors.} The generated explanations could suffer with the same drawbacks as those with existing image captioning models--the sentences being driven more by the priors in the training data and being less grounded in the image. 

As a way of measuring robustness against such priors, we first report the percentage of explanations generated by our model on the validation set that are exact copies from the training set in~\autoref{tbl:sentence_overlap}. While the percentage of duplicates is extremely low for ACT-X, we see a high ratio for VQA-X. To investigate this issue, we measure how the same model trained on descriptions perform. As can be seen in the left two columns of~\autoref{tbl:sentence_overlap}, the percentage is low for both datasets. The VQA-X dataset currently only has 1 explanation per (Img, Q, A) triplet, while ACT-X, MHP descriptions~\cite{RAMTSL16}, and COCO datasets have at least 3 sentences per image. We postulate that our model shows robustness against statistical priors given the training sentences are diverse enough.

As another way of measuring robustness, we investigate whether the generated explanations change across images for a given question and answer pair, and vice versa. The results are detailed in~\autoref{sec:res:qual}.

\begin{table}[tb]
\small
\centering
\begin{tabular}{l|r|r}
\toprule
& VQA-X & ACT-X \\
\midrule
Random Point & 9.21 & 9.36 \\
Uniform & 5.56 &  4.81 \\
Ours (ans-att) & 4.24 & 6.44 \\
Ours (exp-att) & 4.31 & \textbf{3.8} \\
Finetuned (ans-att) & \textbf{4.24} & -- \\
Finetuned (exp-att) & \textbf{4.25} & -- \\
\bottomrule
\end{tabular}

\caption{Evaluation of pointing with Earth Mover's distance (lower is better). Ours (ans-att) denotes the attention map used to predict the answer whereas Ours (exp-att) denotes the attention map used to generate explanations.
}
\label{tbl:attention_map_eval_wo}
\end{table}

\begin{table}[tb]
\small
\centering
\begin{tabular}{l|r|r|r}
\toprule
& VQA-X & ACT-X & VQA-HAT \\
\midrule
Random Point & -0.0010 & +0.0003 & -0.0001 \\
Uniform & -0.0002 & -0.0007 & -0.0007 \\
HieCoAtt-Q~\cite{DBLP:journals/corr/DasAZPB16} & -- & -- & 0.2640 \\
Ours (ans-att) & +0.2280 & +0.0387 & +0.1366 \\
Ours (exp-att) & +0.3132 & \textbf{+0.3744} & +0.3988 \\
Finetuned (ans-att) & +0.2290 & -- & +0.2809 \\
Finetuned (exp-att) & \textbf{+0.3152} & -- & \textbf{+0.5041} \\
\bottomrule
\end{tabular}

\caption{Evaluation of pointing with Rank Correlation metric (higher is better). Ours (ans-att) denotes the attention map used to predict the answer whereas Ours (exp-att) denotes the attention map used to generate explanations. All the results here have a standard error of less than 0.005.
}
\label{tbl:attention_map_eval_rank_corr}
\end{table}

\subsection{Visual Pointing}
\label{sec:res:pointing}
We compare our attention maps to several baselines and report quantitative results with corresponding analysis.

\myparagraph{Details on compared baselines.}
We compare our model against the following baselines. \textit{Random Point} randomly attends to a single point in a $20\times 20$ grid. \textit{Uniform Map} generates attention map that is uniformly distributed over the $20\times 20$ grid. %

\myparagraph{Comparing with baselines.} We evaluate attention maps using the Earth Mover's Distance (lower is better) and rank correlation (higher is better) on VQA-X and ACT-X datasets in~\autoref{tbl:attention_map_eval_wo} and~\autoref{tbl:attention_map_eval_rank_corr}. From ~\autoref{tbl:attention_map_eval_wo}, we observe that our exp-att outperforms  baselines and performs similarly as ans-att for VQA-X, indicating that exp-att not only aligns well with human annotated attentions, but also with the model attention used for making decision. In fact, the EMD and rank correlation between VQA ans-att and exp-att are 3.153 and 0.4563 respectively, indicating high alignment. For ACT-X, our exp-att outperforms all the baselines and the ans-att, indicating that the regions the model attends to when generating an explanation agree more with regions humans point to when justifying a decision. This suggests that whereas ans-att attention maps can be helpful for understanding a model and debugging, they are not necessarily the best option when providing visual evidence which agrees with human justifications.

A direct comparison between our dataset and VQA-HAT dataset from~\cite{DBLP:journals/corr/DasAZPB16} is currently not viable because the two datasets have different splits and the overlap is only 9 QA pairs. However, we instead compute the rank-correlation metric following~\cite{DBLP:journals/corr/DasAZPB16} for their and our datasets. In~\autoref{tbl:attention_map_eval_rank_corr}, we see similar trends as in the EMD metric where our model outperforms the baseline in all datasets and the best model in~\cite{DBLP:journals/corr/DasAZPB16} for the rank-correlation metric.

\begin{figure}[t]
\begin{center}
  \includegraphics[width=.9\linewidth]{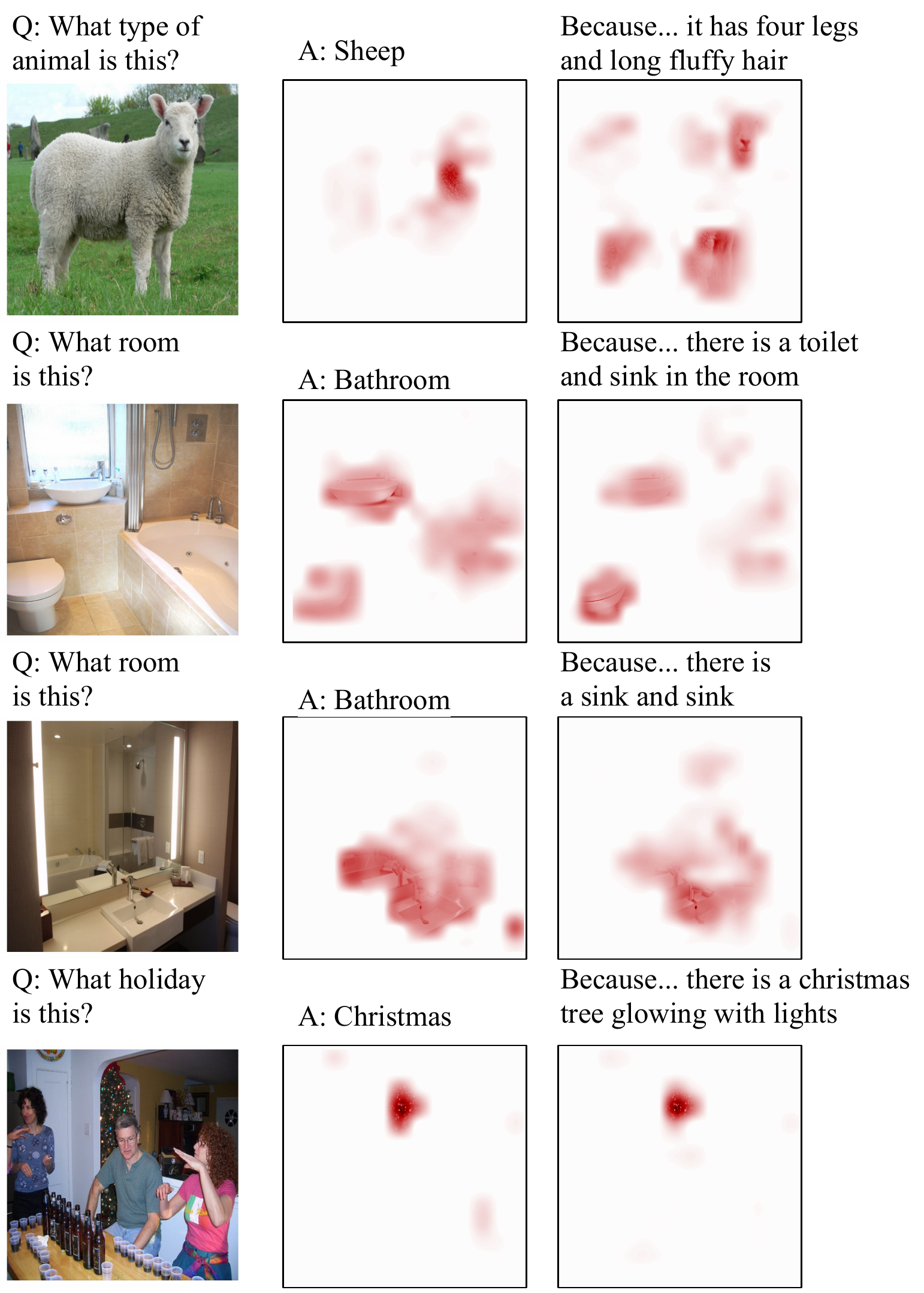}
\end{center}
\caption{VQA-X qualitative results: For the given image and question (column 1), the model provides an answer and the pointing evidence for that answer (column 2), and a justification and the pointing evidence for that justification (column 3).}
\label{fig:VQAqualitative}
\end{figure}

\begin{figure}[t]
\begin{center}
  \includegraphics[width=.9\linewidth]{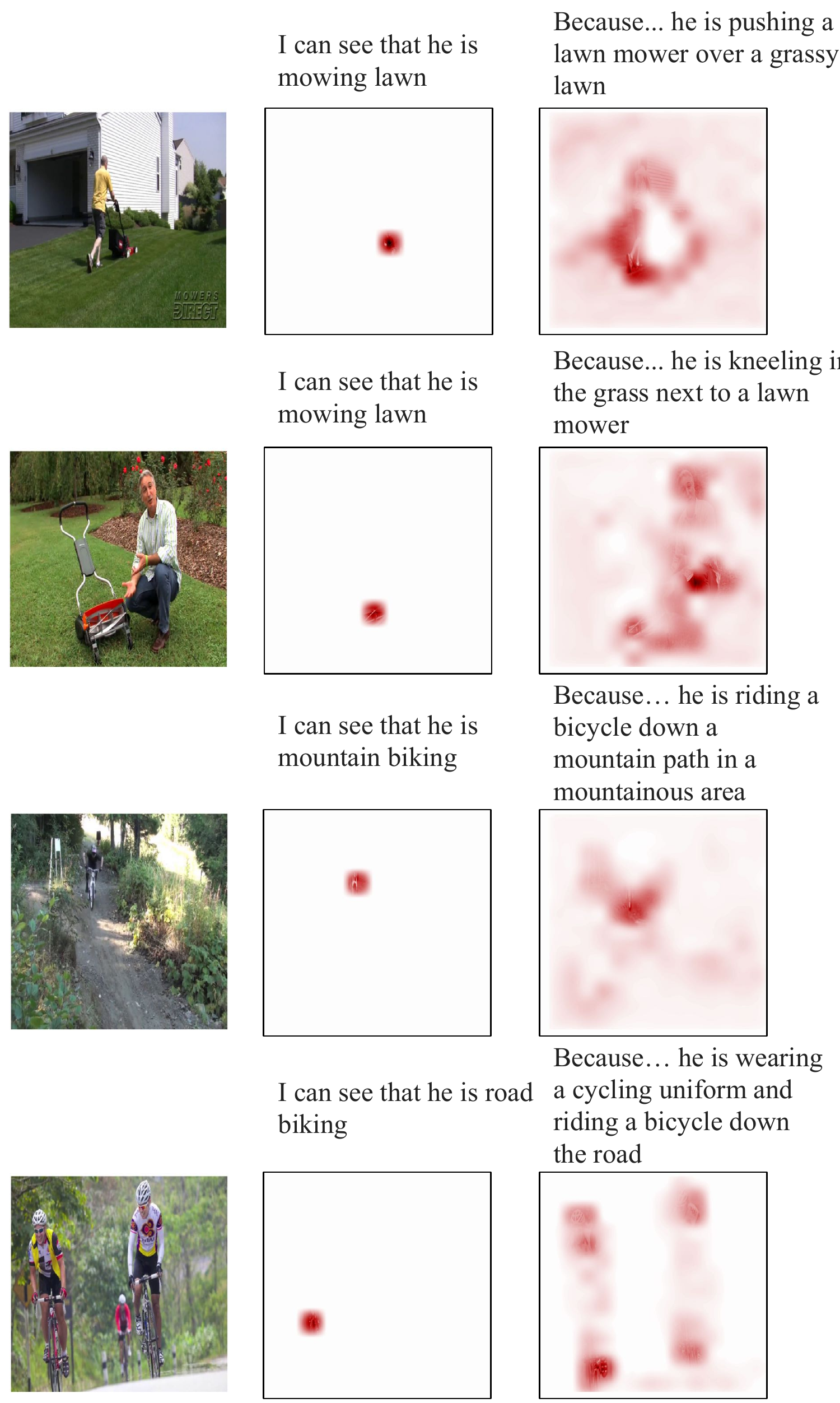}
\end{center}
\caption{ACT-X qualitative results: For the given image (column 1), the model provides an answer and the pointing evidence for that answer (column 2), and a justification and the pointing evidence for that justification (column 3).}
\label{fig:ACTqualitative}
\end{figure}

\subsection{Qualitative Results}
\label{sec:res:qual}
In this section we present our qualitative results on VQA-X and ACT-X datasets demonstrating that our model generates high quality sentences and the attention maps point to relevant locations in the image.

\myparagraph{VQA-X.}  \autoref{fig:VQAqualitative} shows qualitative results on our VQA-X dataset.
Our textual justifications are able to both capture common sense and discuss specific image parts important for answering a question.
For example, when asked what holiday it is, the explanation model is able to discuss what object may represent the concept of ''Christmas'', \ie ``there is a christmas tree glowing with lights.''
When determining the kind of animal which requires discussing specific image parts, the textual justification discusses the legs and the fact that the animal has long fluffy hair.

Visually, we notice that our attention model is able to point to important visual evidence.  
For example in the second row of~\autoref{fig:VQAqualitative}, for the question ``what room is this?'' the visual explanation focuses on the toilet and the sink. 
Given the same QA pair but with different image where there is no toilet, our attention model is able to focus on the sink and its reflection on the mirror.
Moreover, supporting our initial claims, the attention map that leads to the correct answer and the attention map that leads to a relevant explanation may look different, \eg generating ``it has four legs and long fluffy hair'' requires looking at the sheep with a wider angle. 

\myparagraph{ACT-X.} \autoref{fig:ACTqualitative} shows results on our ACT-X dataset.
Textual explanations discuss a variety of visual cues important for correctly classifying activities such as 
global context, \eg ``over a grassy lawn / in a mountainous area'', and person-object interaction, \eg``pushing a lawn mower / riding a bicycle'' for mowing lawn and mountain biking, respectively.
These explanations require determining which of many multiple cues are appropriate to justify a particular action.

\begin{figure*}[h]
\begin{center}
  \includegraphics[width=\linewidth]{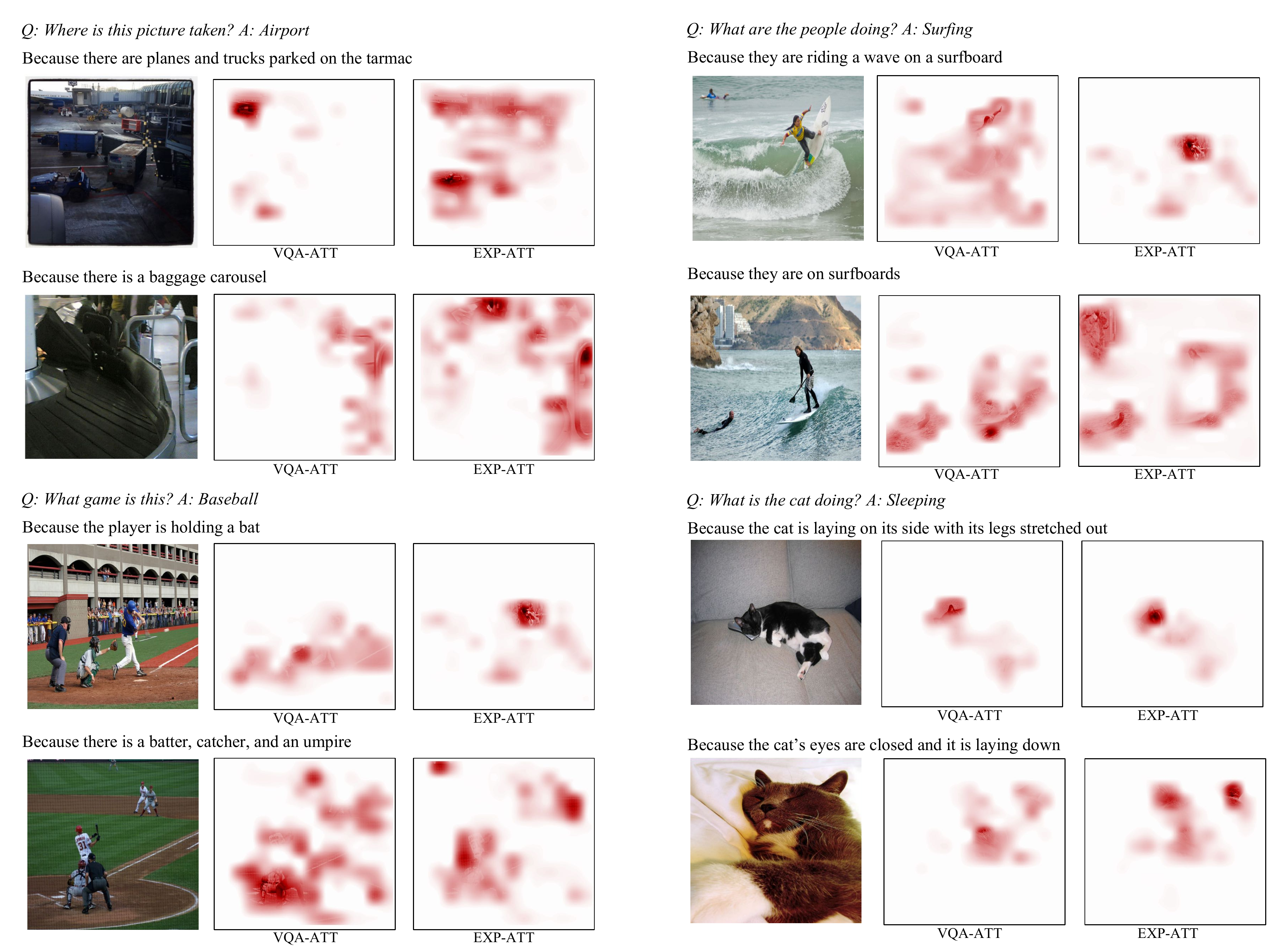}
\end{center}
\caption{VQA-X results with the same question/answer pair. We select results with the same question and answer pair with two different images and show that although the QA pairs are the same, for different images our model generates different explanations (Answers are correctly predicted). VQA-ATT denotes attention maps used for predicting answers and EXP-ATT denotes attention maps used for generating the corresponding justifications.}
\label{fig:VQAqualitative-sameQA}
\end{figure*}

\begin{figure*}[h]
\begin{center}
  \includegraphics[width=\linewidth]{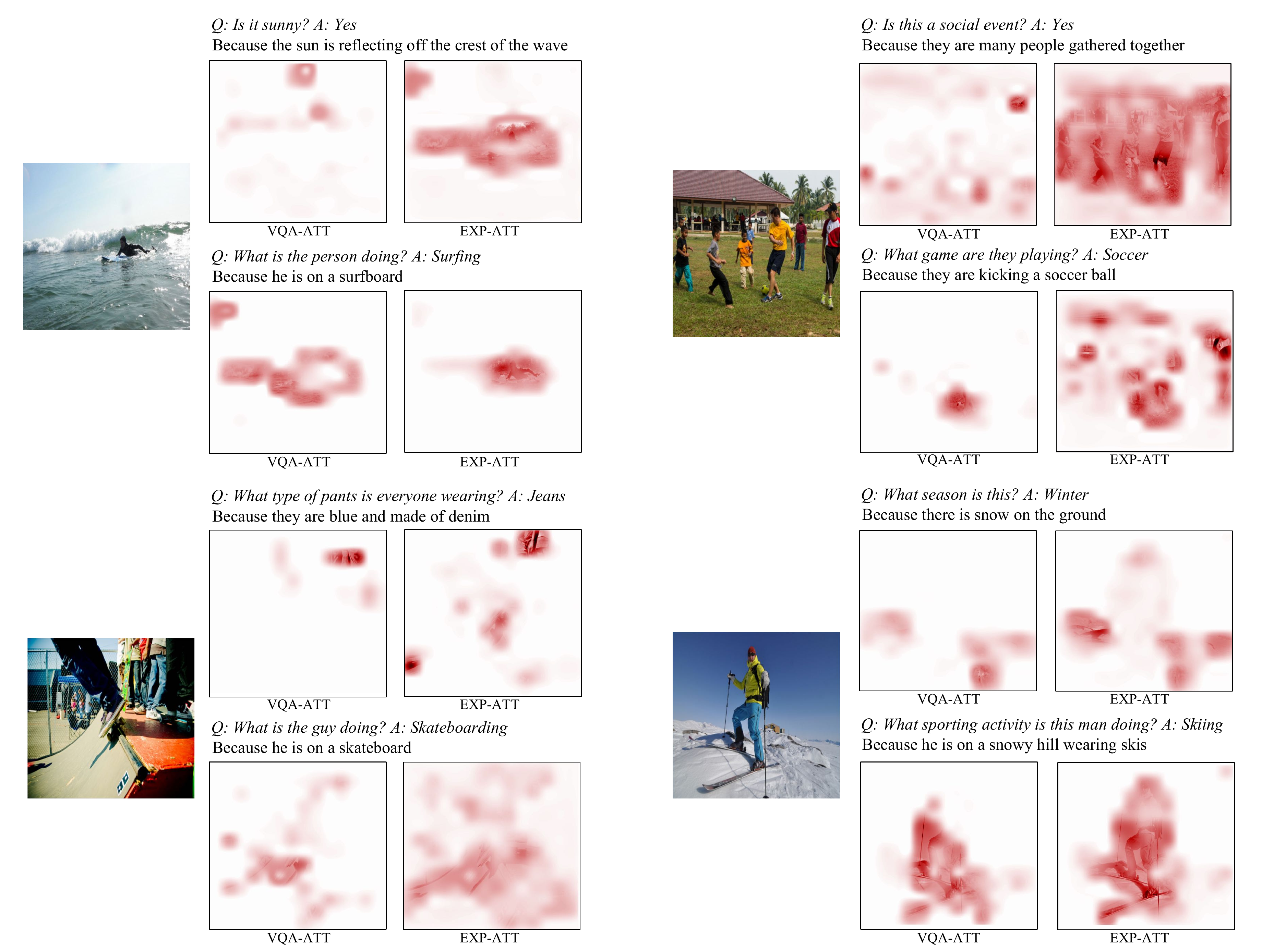}
\end{center}
\caption{VQA-X results with same image and different questions. We select results with the same image and different Q/A pairs and show that although the images are the same, our model is able to answer the questions differently and generate a different explanation accordingly (Answers are correctly predicted). VQA-ATT denotes attention maps used for predicting answers and EXP-ATT denotes attention maps used for generating the corresponding justifications.}
\label{fig:VQAqualitative-sameimage}
\end{figure*}

Our model points to visual evidence important for understanding each human activity.
For example to classify ``mowing lawn'' in the second row of ~\autoref{fig:ACTqualitative} the model focuses both on the person, who is on the grass, as well as the lawn mower. 
Our model can also differentiate between similar activities based off of context, \eg''mountain biking'' or ''road biking''.
\myparagraph{Additional Results in Various Settings.}
\autoref{fig:VQAqualitative-sameQA} 
and \autoref{fig:VQAqualitative-sameimage} 
demonstrate that both images and the question/answer pair are needed for good explanations. They also demonstrate that the explanations generated by our model are visually grounded and are robust to priors existing in the training data.

\autoref{fig:VQAqualitative-sameQA} shows explanations for different images, but with the same question/answer pair.
Importantly, explanation text and visualizations change to reflect image content. For instance, for the question ''Where is this picture taken?'' our model explains the answer ''Airport'' by pointing and discussing planes and trucks in the first image while pointing and discussing baggage carousel in the second image.

\autoref{fig:VQAqualitative-sameimage} shows that when different questions are asked about the same images, explanations provide information which are specific to the questions. For example, for the question ''Is it sunny?'' our model explains the answer ''Yes'' by mentioning the sun and its reflection and pointing to the sky and the water, whereas for the question ''What is the person doing?'' it points more directly to the surfer and mentions that the person is on a surfboard. 

\begin{figure*}[h]
\begin{center}
 \includegraphics[width=\linewidth]{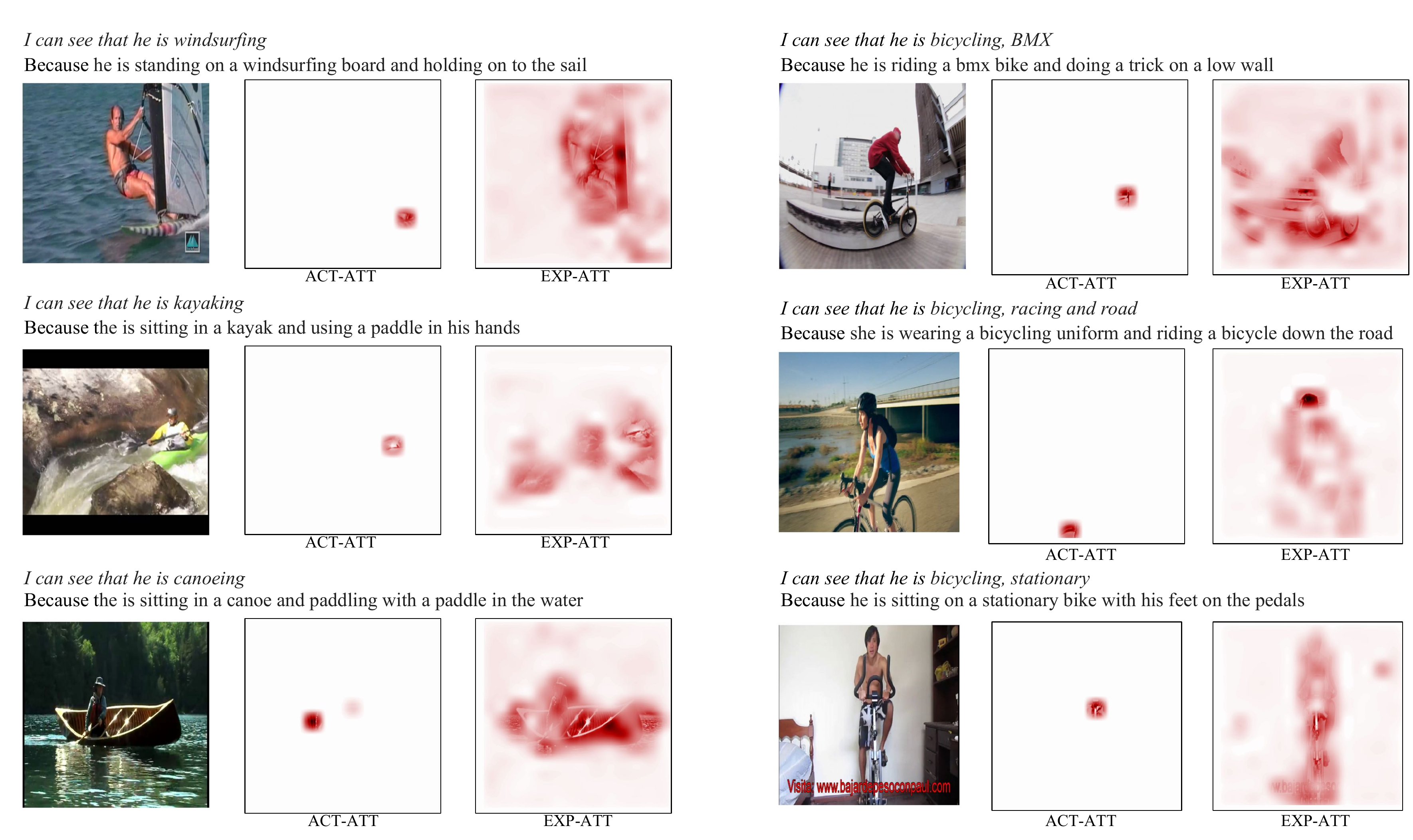}
\end{center}
\caption{ACT-X results with similar activities. Figure on the left: We show results with fine-grained activities all related to windsurfing, kayaking, canoeing and observe that both the fine-grained activities are correctly predicted and the explanations match the activity and the image. Figure on the right: We show results with fine-grained activities all related to bicycling and observe that both the fine-grained activities are correctly predicted and the explanations match the activity and the image. ACT-ATT denotes attention maps used for predicting answers and EXP-ATT denotes attention maps used for generating the corresponding justifications.}
\label{fig:ACTqualitative:similar:activity}
\end{figure*}

\autoref{fig:ACTqualitative:similar:activity} shows that explanations on the ACT-X dataset discuss small details important for differentiating between similar classes.  For example, when explaining kayaking and windsurfing, it is important to mention the correct sporting equipment such as ''kayak'' and ''sail'' instead of image context. On the other hand, when distinguishing bicycling (BMX) and bicycling (racing and road), it is important to discuss the image context such as ''doing a trick on a low wall'' and ''riding a bicycle down the road.''

\begin{figure*}[h]
\begin{center}
 \includegraphics[width=\linewidth]{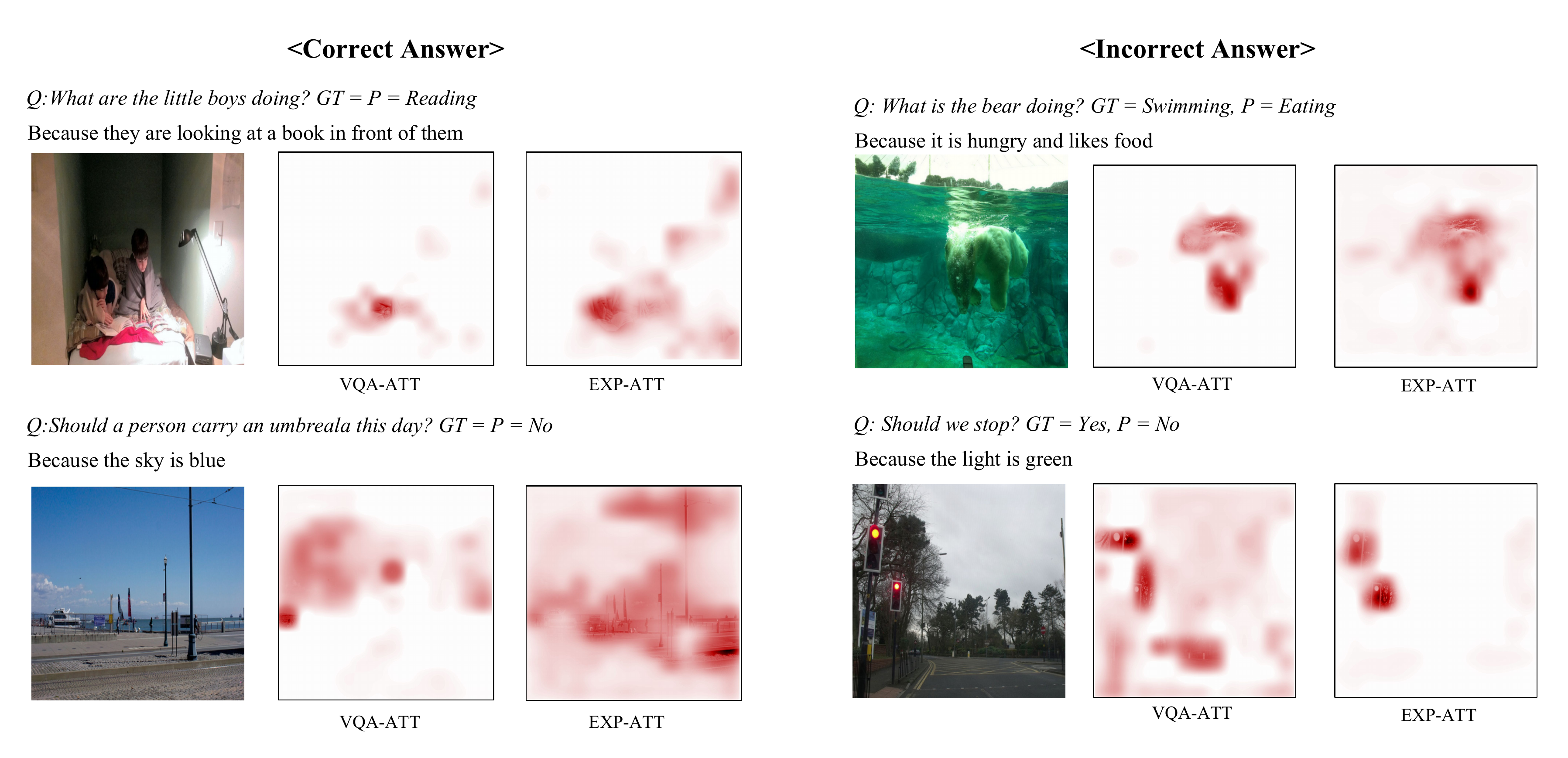}
\end{center}
\caption{VQA-X results. GT denotes ground-truth answer while P indicates actual prediction made by the model. Figure on the left: We show various qualitative results with correctly predicted answer and observe that the explanation justifies the answer accordingly. Figure on the right: We show results with incorrectly predicted answer and observe that although the answer is incorrect, our model can provide visual and textual explanations on why the model might be failing in those cases.}
\label{fig:VQAqualitative:correct_incorrect}
\end{figure*}

\begin{figure*}[h]
\begin{center}
 \includegraphics[width=\linewidth]{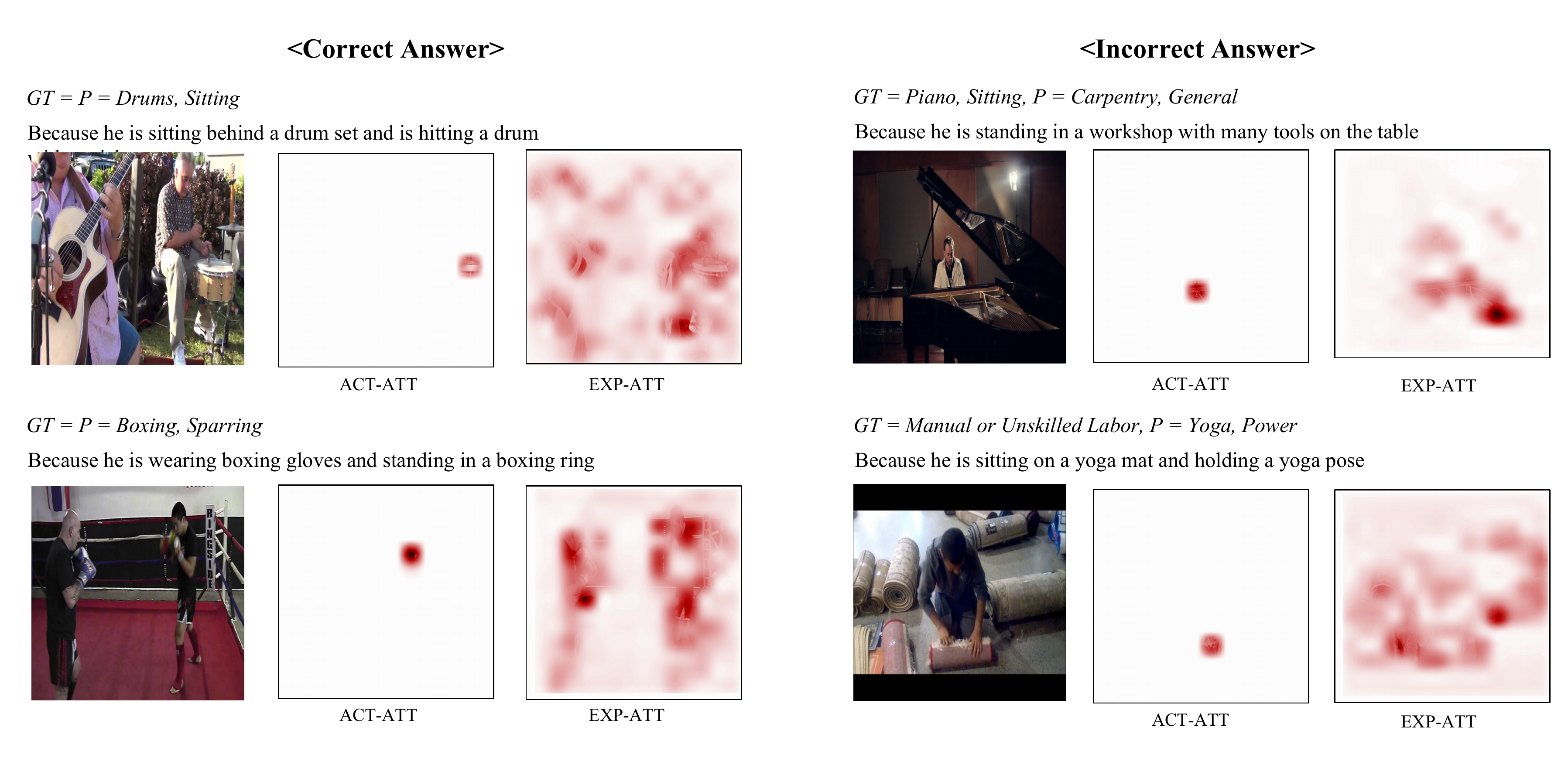}
\end{center}
\caption{ACT-X results. GT denotes ground-truth answer while P indicates actual prediction made by the model. Figure on the left: We show various qualitative results with correctly predicted answer and observe that the explanation justifies the answer accordingly. Figure on the right: We show results with incorrectly predicted answer and observe that although the answer is incorrect, our model can provide visual and textual explanations on why the model might be failing in those cases.}
\label{fig:ACTqualitative:correct_incorrect}
\end{figure*}

\autoref{fig:VQAqualitative:correct_incorrect} and \autoref{fig:ACTqualitative:correct_incorrect} compare explanations when the answers or action labels are correctly and incorrectly predicted. In addition to providing an intuition about why predictions are correct, our explanations frequently justify why the model makes incorrect predictions. For example, when incorrectly predicting whether one should stop or go (~\autoref{fig:VQAqualitative:correct_incorrect}, lower-right example), the model outputs ''Because the light is green'' suggesting that the model has mistaken a red light for a green light, and furthermore, that green lights mean ''go''.

~\autoref{fig:ACTqualitative:correct_incorrect} shows similar trends on the ACT-X dataset. For example, when incorrectly predicting the activity power yoga for an image depicting manual labor, the explanation ''Because he is sitting on a yoga mat and holding a yoga pose” suggests that the rug may have been misclassified as a yoga mat. We reiterate that our model justifies predictions and does not fully explain the inner-workings of deep architectures. However, these justifications demonstrate that our model can output intuitive explanations which could help those unfamiliar with deep architectures make sense of model predictions.
\section{Conclusion}
\label{sec:conc}
As a step towards explainable AI models, in this work we introduced a novel attentive explanation model that is capable of providing natural language justifications of decisions as well as pointing to the evidence. We proposed two novel explanation datasets collected through crowd sourcing for visual question answering and activity recognition, \ie VQA-X and ACT-X. We quantitatively demonstrated that both attention and using reference explanations to train our model helps achieve high quality explanations. Furthermore, we demonstrated that our model is able to point to the evidence as well as to give natural sentence justifications, similar to ones humans give. %

\section*{Acknowledgements}
This work was in part supported by DARPA; AFRL; DoD MURI award N000141110688; NSF awards  IIS-1212798, IIS-1427425, and IIS-1536003, and the Berkeley Artificial Intelligence Research (BAIR) Lab.
\newpage

\newpage

\newpage

\newpage

\newpage

\clearpage
\bibliographystyle{ieee}
\bibliography{biblioLong,biblio,egbib}

\begin{thebibliography}{10}\itemsep=-1pt

\bibitem{spice2016}
P.~Anderson, B.~Fernando, M.~Johnson, and S.~Gould.
\newblock Spice: Semantic propositional image caption evaluation.
\newblock In {\em Proceedings of the European Conference on Computer Vision
  (ECCV)}, 2016.

\bibitem{APGS14}
M.~Andriluka, L.~Pishchulin, P.~Gehler, and B.~Schiele.
\newblock 2d human pose estimation: New benchmark and state of the art
  analysis.
\newblock In {\em Proceedings of the IEEE Conference on Computer Vision and
  Pattern Recognition (CVPR)}, 2014.

\bibitem{antol2015vqa}
S.~Antol, A.~Agrawal, J.~Lu, M.~Mitchell, D.~Batra, C.~L. Zitnick, and
  D.~Parikh.
\newblock Vqa: Visual question answering.
\newblock In {\em Proceedings of the IEEE International Conference on Computer
  Vision (ICCV)}, 2015.

\bibitem{bahdanau2014neural}
D.~Bahdanau, K.~Cho, and Y.~Bengio.
\newblock Neural machine translation by jointly learning to align and
  translate.
\newblock In {\em Proceedings of the International Conference on Learning
  Representations (ICLR)}, 2015.

\bibitem{banerjee2005meteor}
S.~Banerjee and A.~Lavie.
\newblock Meteor: An automatic metric for mt evaluation with improved
  correlation with human judgments.
\newblock In {\em Proceedings of the ACL Workshop on Intrinsic and Extrinsic
  Evaluation Measures for Machine Translation and/or Summarization}, volume~29,
  pages 65--72, 2005.

\bibitem{bell13opensurfaces}
S.~Bell, P.~Upchurch, N.~Snavely, and K.~Bala.
\newblock Opensurfaces: A richly annotated catalog of surface appearance.
\newblock In {\em SIGGRAPH Conf. Proc.}, volume~32, 2013.

\bibitem{berg2013you}
T.~Berg and P.~N. Belhumeur.
\newblock How do you tell a blackbird from a crow?
\newblock In {\em Proceedings of the IEEE International Conference on Computer
  Vision (ICCV)}, 2013.

\bibitem{biran2014justification}
O.~Biran and K.~McKeown.
\newblock Justification narratives for individual classifications.
\newblock In {\em Proceedings of the AutoML workshop at ICML}, volume 2014,
  2014.

\bibitem{chen2015abc}
K.~Chen, J.~Wang, L.-C. Chen, H.~Gao, W.~Xu, and R.~Nevatia.
\newblock Abc-cnn: An attention based convolutional neural network for visual
  question answering.
\newblock {\em arXiv:1511.05960}, 2015.

\bibitem{core2006building}
M.~G. Core, H.~C. Lane, M.~Van~Lent, D.~Gomboc, S.~Solomon, and M.~Rosenberg.
\newblock Building explainable artificial intelligence systems.
\newblock In {\em Proceedings of the national conference on artificial
  intelligence}, volume~21, page 1766. Menlo Park, CA; Cambridge, MA; London;
  AAAI Press; MIT Press; 1999, 2006.

\bibitem{DBLP:journals/corr/DasAZPB16}
A.~Das, H.~Agrawal, C.~L. Zitnick, D.~Parikh, and D.~Batra.
\newblock Human attention in visual question answering: Do humans and deep
  networks look at the same regions?
\newblock {\em CoRR}, abs/1606.03556, 2016.

\bibitem{doersch2012makes}
C.~Doersch, S.~Singh, A.~Gupta, J.~Sivic, and A.~Efros.
\newblock What makes paris look like paris?
\newblock {\em ACM Transactions on Graphics}, 31(4), 2012.

\bibitem{donahue16pami}
J.~Donahue, L.~A. Hendricks, M.~Rohrbach, S.~Venugopalan, S.~Guadarrama,
  K.~Saenko, and T.~Darrell.
\newblock Long-term recurrent convolutional networks for visual recognition and
  description.
\newblock {\em IEEE Transactions on Pattern Analysis and Machine Intelligence
  (TPAMI)}, 2016.

\bibitem{escorcia2015relationship}
V.~Escorcia, J.~C. Niebles, and B.~Ghanem.
\newblock On the relationship between visual attributes and convolutional
  networks.
\newblock In {\em Proceedings of the IEEE Conference on Computer Vision and
  Pattern Recognition (CVPR)}, 2015.

\bibitem{fukui16emnlp}
A.~Fukui, D.~H. Park, D.~Yang, A.~Rohrbach, T.~Darrell, and M.~Rohrbach.
\newblock Multimodal compact bilinear pooling for visual question answering and
  visual grounding.
\newblock In {\em Proceedings of the Conference on Empirical Methods in Natural
  Language Processing (EMNLP)}, 2016.

\bibitem{gao16cvpr}
Y.~Gao, O.~Beijbom, N.~Zhang, and T.~Darrell.
\newblock Compact bilinear pooling.
\newblock In {\em Proceedings of the IEEE Conference on Computer Vision and
  Pattern Recognition (CVPR)}, 2016.

\bibitem{hendricks16eccv}
L.~A. Hendricks, Z.~Akata, M.~Rohrbach, J.~Donahue, B.~Schiele, and T.~Darrell.
\newblock Generating visual explanations.
\newblock In {\em Proceedings of the European Conference on Computer Vision
  (ECCV)}, 2016.

\bibitem{DBLP:journals/corr/KimOKHZ16}
J.~Kim, K.~W. On, J.~Kim, J.~Ha, and B.~Zhang.
\newblock Hadamard product for low-rank bilinear pooling.
\newblock {\em CoRR}, abs/1610.04325, 2016.

\bibitem{lane2005explainable}
H.~C. Lane, M.~G. Core, M.~Van~Lent, S.~Solomon, and D.~Gomboc.
\newblock Explainable artificial intelligence for training and tutoring.
\newblock Technical report, DTIC Document, 2005.

\bibitem{L04}
C.-Y. Lin.
\newblock Rouge: a package for automatic evaluation of summaries.
\newblock In {\em Text Summarization Branches Out: Proceedings of the ACL-04
  Workshop}, 2004.

\bibitem{lin14eccv}
T.-Y. Lin, M.~Maire, S.~Belongie, J.~Hays, P.~Perona, D.~Ramanan,
  P.~Doll{\'a}r, and C.~L. Zitnick.
\newblock Microsoft coco: Common objects in context.
\newblock In {\em Proceedings of the European Conference on Computer Vision
  (ECCV)}, pages 740--755. Springer, 2014.

\bibitem{lin2014microsoft}
T.-Y. Lin, M.~Maire, S.~Belongie, J.~Hays, P.~Perona, D.~Ramanan,
  P.~Doll{\'a}r, and C.~L. Zitnick.
\newblock Microsoft coco: Common objects in context.
\newblock In {\em Proceedings of the European Conference on Computer Vision
  (ECCV)}, 2014.

\bibitem{malinowski15iccv}
M.~Malinowski, M.~Rohrbach, and M.~Fritz.
\newblock Ask your neurons: A neural-based approach to answering questions
  about images.
\newblock In {\em Proceedings of the IEEE International Conference on Computer
  Vision (ICCV)}, 2015.

\bibitem{mallya16eccv}
A.~Mallya and S.~Lazebnik.
\newblock Learning models for actions and person-object interactions with
  transfer to question answering.
\newblock In {\em Proceedings of the European Conference on Computer Vision
  (ECCV)}, 2016.

\bibitem{papineni2002bleu}
K.~Papineni, S.~Roukos, T.~Ward, and W.-J. Zhu.
\newblock Bleu: a method for automatic evaluation of machine translation.
\newblock In {\em Proceedings of the Annual Meeting of the Association for
  Computational Linguistics (ACL)}, pages 311--318, 2002.

\bibitem{pele2009}
O.~Pele and M.~Werman.
\newblock Fast and robust earth mover's distances.
\newblock In {\em 2009 IEEE 12th International Conference on Computer Vision},
  pages 460--467. IEEE, September 2009.

\bibitem{pishchulin14gcpr}
L.~Pishchulin, M.~Andriluka, and B.~Schiele.
\newblock Fine-grained activity recognition with holistic and pose based
  features.
\newblock In {\em Proceedings of the German Confeence on Pattern Recognition
  (GCPR)}, pages 678--689. Springer, 2014.

\bibitem{RALS16}
S.~Reed, Z.~Akata, H.~Lee, and B.~Schiele.
\newblock Learning deep representations of fine-grained visual descriptions.
\newblock In {\em Proceedings of the IEEE Conference on Computer Vision and
  Pattern Recognition (CVPR)}, 2016.

\bibitem{RAMTSL16}
S.~Reed, Z.~Akata, S.~Mohan, S.~Tenka, B.~Schiele, and H.~Lee.
\newblock Learning what and where to draw.
\newblock In {\em Advances in Neural Information Processing Systems (NIPS)},
  2016.

\bibitem{RubnerTG98iccv}
Y.~Rubner, C.~Tomasi, and L.~J. Guibas.
\newblock A metric for distributions with applications to image databases.
\newblock In {\em Proceedings of the IEEE International Conference on Computer
  Vision (ICCV)}, 1998.

\bibitem{shih2015look}
K.~J. Shih, S.~Singh, and D.~Hoiem.
\newblock Where to look: Focus regions for visual question answering.
\newblock In {\em Proceedings of the IEEE Conference on Computer Vision and
  Pattern Recognition (CVPR)}, 2016.

\bibitem{shortliffe1975model}
E.~H. Shortliffe and B.~G. Buchanan.
\newblock A model of inexact reasoning in medicine.
\newblock {\em Mathematical biosciences}, 23(3):351--379, 1975.

\bibitem{van2004explainable}
M.~Van~Lent, W.~Fisher, and M.~Mancuso.
\newblock An explainable artificial intelligence system for small-unit tactical
  behavior.
\newblock In {\em NCAI}, 2004.

\bibitem{vedantam2015cider}
R.~Vedantam, C.~Lawrence~Zitnick, and D.~Parikh.
\newblock Cider: Consensus-based image description evaluation.
\newblock In {\em Proceedings of the IEEE Conference on Computer Vision and
  Pattern Recognition (CVPR)}, pages 4566--4575, 2015.

\bibitem{CaltechUCSDBirdsDataset}
P.~Welinder, S.~Branson, T.~Mita, C.~Wah, F.~Schroff, S.~Belongie, and
  P.~Perona.
\newblock {Caltech-UCSD Birds 200}.
\newblock Technical Report CNS-TR-2010-001, Caltech, 2010.

\bibitem{wick1992reconstructive}
M.~R. Wick and W.~B. Thompson.
\newblock Reconstructive expert system explanation.
\newblock {\em Artificial Intelligence}, 54(1-2):33--70, 1992.

\bibitem{xiong16dynamic}
C.~Xiong, S.~Merity, and R.~Socher.
\newblock Dynamic memory networks for visual and textual question answering.
\newblock In {\em Proceedings of the International Conference on Machine
  Learning (ICML)}, 2016.

\bibitem{xu2015ask}
H.~Xu and K.~Saenko.
\newblock Ask, attend and answer: Exploring question-guided spatial attention
  for visual question answering.
\newblock In {\em Proceedings of the European Conference on Computer Vision
  (ECCV)}, 2016.

\bibitem{yang2015stacked}
Z.~Yang, X.~He, J.~Gao, L.~Deng, and A.~Smola.
\newblock Stacked attention networks for image question answering.
\newblock In {\em Proceedings of the IEEE Conference on Computer Vision and
  Pattern Recognition (CVPR)}, 2016.

\bibitem{zeiler2014visualizing}
M.~D. Zeiler and R.~Fergus.
\newblock Visualizing and understanding convolutional networks.
\newblock In {\em Proceedings of the European Conference on Computer Vision
  (ECCV)}. 2014.

\bibitem{zhou2014object}
B.~Zhou, A.~Khosla, A.~Lapedriza, A.~Oliva, and A.~Torralba.
\newblock Object detectors emerge in deep scene cnns.
\newblock In {\em Proceedings of the International Conference on Learning
  Representations (ICLR)}, 2015.

\bibitem{zhu16cvpr}
Y.~Zhu, O.~Groth, M.~Bernstein, and L.~Fei-Fei.
\newblock {Visual7W: Grounded Question Answering in Images}.
\newblock In {\em Proceedings of the IEEE Conference on Computer Vision and
  Pattern Recognition (CVPR)}, 2016.

\end{thebibliography}

\end{document}